\documentclass[twoside,11pt]{article}

\usepackage{blindtext}

\usepackage{jmlr2e}

\usepackage{lastpage}
\jmlrheading{27}{2026}{1-\pageref{LastPage}}{5/23; Revised
9/25}{4/26}{23-0582}{Yikun Ban, Yuchen Yan, Arindam Banerjee, Jingrui He}

\ShortHeadings{Neural Exploitation and Exploration of Contextual Bandits}{Yikun Ban, Yuchen Yan, Arindam Banerjee, Jingrui He}
\firstpageno{1}

\ShortHeadings{Neural Exploitation and Exploration of Contextual Bandits}{Ban, Yan, Banerjee, and He}
\firstpageno{1}

\usepackage{amsmath, amssymb} 
\usepackage{hyperref}
\usepackage[utf8]{inputenc}
\usepackage[T1]{fontenc}    
\usepackage{amsfonts}       
\usepackage{nicefrac}       
\usepackage{microtype}      
\usepackage{xcolor}         
\usepackage{natbib}
\usepackage{algorithm, algpseudocode} %
\renewcommand{\algorithmicrequire}{\textbf{Input:}}
\renewcommand{\algorithmicensure}{\textbf{Output:}}
\usepackage{enumerate, paralist}
\usepackage{natbib}
\usepackage{comment}
\usepackage{multirow}
\usepackage{xspace}
\usepackage{rotating}
\usepackage{graphicx}
\usepackage{booktabs} 
\usepackage{tabularx}
\usepackage{enumitem}
\usepackage{bbm}
\usepackage{thmtools} 
\usepackage{thm-restate}
\usepackage{caption}
\usepackage{subcaption}
\hypersetup{hidelinks}

\newtheorem{assumption}{Assumption}[section]

\numberwithin{equation}{section}

\newcommand{\bD}{\mathbf{D}}
\newcommand{\bh}{\mathbf{h}}

\newcommand{\deter}{\text{\normalfont det}}

\newcommand{\bxt}{\mathbf{x}_t}
\newcommand{\btheta}{\boldsymbol \theta}
\newcommand{\bx}{\mathbf{x}}
\newcommand{\bw}{\mathbf{W}}
\newcommand{\bbr}{\mathbb{R}}
\newcommand{\bbe}{\mathbb{E}}
\newcommand{\bbp}{\mathbb{P}}

\newcommand{\sysn}{EE-Net\xspace}
\newcommand{\para}[1]{\noindent \textbf{#1}\xspace}
\newcommand{\cald}{\mathcal{D}}
\newcommand{\calb}{\mathcal{B}}
\newcommand{\calo}{\mathcal{O}}
\newcommand{\calh}{\mathcal{H}}
\newcommand{\hide}[1]{}
\newcommand{\tri}{\triangledown}
\newcommand{\hi}{\hat{i}}

\newcommand{\call}{\mathcal{L}}
\newcommand{\iast}{i^{\ast}}
\newcommand{\wcalo}{\widetilde{\calo}}
\newcommand{\hd}{\widetilde{d}}

\newcommand{\htheta}{\widehat{\btheta}}

\newcommand{\bsg}{\mathbf{G}}

\newcommand{\bsr}{\mathbf{r}}

\newcommand{\hx}{h(\mathbf{x}_t)}

\newcommand{\gx}{g(\mathbf{x}_t; \btheta_0)}
\newcommand{\bts}{\boldsymbol{\theta}^{\ast}}

\usepackage[]{color-edits}
\addauthor{ab}{blue}
\addauthor{abb}{red}
\addauthor{abcomment}{blue}

\begin{document}

\title{
Neural Exploitation and Exploration of Contextual Bandits
}

\author{\name Yikun Ban \email yikunb2@illinois.edu \\
University of Illinois at Urbana-Champaign
\AND
\name Yuchen Yan \email yucheny5@illinois.edu\\
University of Illinois at Urbana-Champaign
\AND
\name Arindam Banerjee \email arindamb@illinois.edu\\
University of Illinois at Urbana-Champaign
\AND
\name Jingrui He  \email jingrui@illinois.edu\\
University of Illinois at Urbana-Champaign
}

\editor{Chris Wiggins}

\maketitle

\begin{abstract} 
In this paper, we study the neural exploration strategy for contextual bandits.
The dilemma of exploitation and exploration widely exists in real-world applications such as recommender systems, online advertising, and clinical trials.
Contextual bandits provide principled methods to solve this dilemma, including two prevalent techniques: Thompson Sampling (TS), and Upper Confidence Bound (UCB).
Neural contextual bandits have been studied to adapt to the non-linear reward function, combined with TS or UCB strategies for exploration.
In this paper, we introduce, EE-Net, which is a novel framework to utilize another neural network to learn the potential gain of exploitation neural network for exploration, different from UCB-based and TS-based approaches that rely on the large-deviation-based statistical confidence bound.
In addition, we provide an instance-based $\widetilde{\mathcal{O}}(\sqrt{T})$ regret upper bound for EE-Net with a new proof workflow. Empirically, we show that EE-Net outperforms related linear and neural contextual bandit baselines on real-world datasets.

\end{abstract}


\section{Introduction}

The stochastic contextual multi-armed bandit (MAB) \citep{lattimore2020bandit} has been extensively studied in the machine learning community for decades as a solution to sequential decision-making problems. This framework has practical applications in various fields, such as online advertising \citep{2010contextual} and personalized recommendation \citep{ban2024neuralp,ban2021local}. In the standard contextual bandit setting, a learner is presented with a set of \( n \) arms in each round, where each arm is characterized by a context vector. The learner then selects and plays one arm based on a specific strategy, receiving a corresponding reward. The objective is to maximize the cumulative rewards over \( T \) rounds.

MAB provides principled solutions for the trade-off between Exploitation and Exploration (EE).
On one hand, the learner should exploit the information from the collected data; on the other hand, the learner should explore the under-explored arms to obtain new information. 
The widely-used strategies for EE trade-off includes the following three techniques: Epsilon-greedy \citep{langford2008epoch}, Thompson Sampling (TS) \citep{thompson1933likelihood, kveton2021meta}, and Upper Confidence Bound (UCB) \citep{auer2002using, ban2020generic}.
Based on the realizability assumptions on reward function, the first mainstream is the linear contextual bandits, where the reward is assumed to be a linear function with respect to arm context vectors. Linear bandits have been well studied and succeeded both empirically and theoretically \citep{bouneffouf2020survey,slivkins2019introduction}.
UCB-based algorithms \citep{2010contextual, chu2011contextual, wu2016contextual,ban2021local} calculate the confidence ellipsoid of estimated reward based on ridge regression. The selection criterion is to pull the arm with the maximal upper confidence bound concerning the estimated reward.
TS-based algorithms \citep{agrawal2013thompson, abeille2017linear, osband2023approximate} formulate each arm with a prior distribution and select the arm with the maximal posterior probability of being the best arm \citep{valko2013finite}.
Another mainstream, neural contextual bandit, has gained attention in recent years.  Neural bandits are able to utilize deep neural networks to learn the underlying non-linear reward functions, thanks to the powerful representation ability.
Considering the past selected arms and received rewards as training samples, a neural network is built for exploitation.
\citet{zhou2020neural} compute a gradient-based upper confidence bound and use the UCB strategy to select arms. \cite{zhang2020neural} formulate each arm as a normal distribution and the standard deviation is calculated based on the gradient of exploitation neural network, and then uses the TS strategy to choose arms.

\begin{figure}[t] 
    \vspace{-1em}
    \includegraphics[width=0.9\columnwidth]{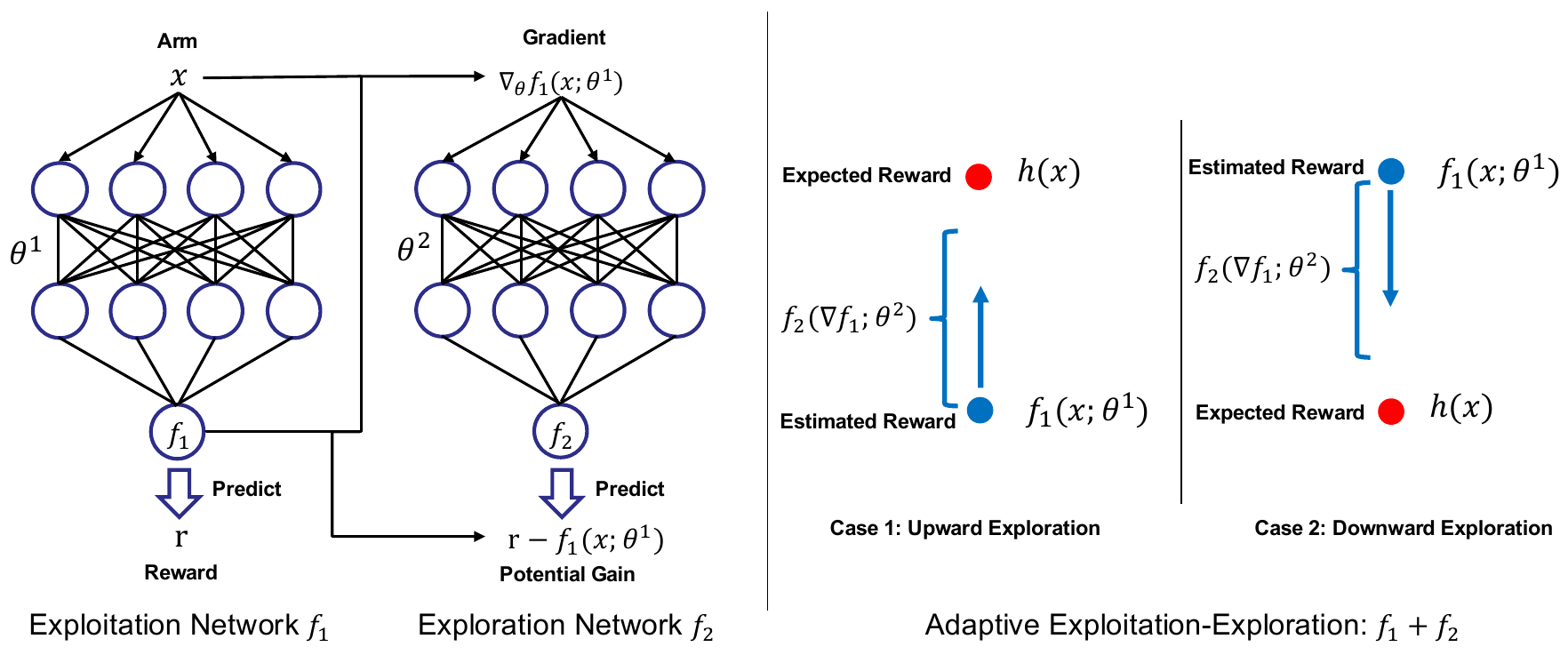}
    \centering
    \caption{ Exploration direction (right side): (1) "Upward" exploration should be performed when the model underestimates the arm's reward; (2) "Downward" exploration should be performed when the model overestimates the arm's reward. 
EE-Net (the proposed strategy), depicted in the left side, intends to adaptively make exploration according to the estimated potential gain of arm. }
       \vspace{-1em}
       \label{fig:structure}
\end{figure}

Figure \ref{fig:structure} illustrates the primary motivation for this work by examining the exploration mechanism in detail. Let \( h(\bx) \) represent the expected reward of an arm \( \bx \), and let \( f_1(\bx; \btheta^1) \) denote the reward estimated by the exploitation model \( f_1 \). The difference between the estimated reward and the expected reward divides exploration into two categories: "upward" exploration and "downward" exploration.
Upward exploration occurs when the expected reward is greater than the estimated reward, i.e., \( h(\bx) > f_1(\bx; \btheta^1) \). This situation indicates that the learner's model has underestimated the reward, necessitating an additional value to be added to the estimated reward to reduce the discrepancy between the expected and estimated rewards. Conversely, downward exploration happens when the expected reward is less than the estimated reward, i.e., \( h(\bx) < f_1(\bx; \btheta^1) \). This scenario suggests that the model has overestimated the reward, thus requiring a value to be subtracted from the estimated reward to minimize the gap.

However, existing exploration strategies may not effectively handle both types of exploration. Specifically, UCB-based approaches (e.g., \citep{zhou2020neural}) select an arm based on the estimated reward \( f_1(\bx; \btheta^1) \) plus its confidence bound radius (a positive value). As a result, these approaches may perform poorly when \( f_1(\bx; \btheta^1) \) overestimates \( h(\bx) \) (downward exploration required). On the other hand, TS-based methods (e.g., \cite{wang2021neural}) select arms based on a reward sampled from a distribution where the mean is \( f_1(\bx; \btheta^1) \). However, this distribution is not dynamically adapted for upward or downward exploration. Finally, epsilon-greedy algorithms (e.g., \cite{auer2002using}) randomly select arms with a certain probability, making them unable to directly adjust to the upward or downward exploration.

In this paper, we propose a novel neural exploration strategy named "EE-Net", to adaptively perform upward and downward exploration. Similar to other neural bandits, EE-Net has an exploitation network $f_1$ to estimate the reward for each arm by exploiting the collected data. The crucial difference from existing works is that EE-Net has an exploration network $f_2$ that leverages the underlying exploitation information of $f_1$ to learn the "potential gain" of an arm compared to its current estimated reward. 
The input of the exploration network $f_2$ is the gradient of $f_1$ to incorporate the information of arm contexts and the discriminative ability of $f_1$. The ground truth for training $f_2$ is the potential gain, which is the residual between the observed reward and the estimated reward by $f_1$.
This potential gain effectively indicates the direction for exploration, guiding $f_2$ to make upward or downward exploration. Ultimately, the two neural networks, $f_1$ and $f_2$, work together to select the arms.
Figure \ref{fig:structure} depicts the architecture of \sysn. To sum up, the contributions of this paper can be summarized as follows:
\begin{enumerate}
    \item We propose a novel neural exploration strategy in contextual bandits, EE-Net, where another neural network is assigned to learn the potential gain compared to the current reward estimate for adaptive exploration, in addition to the neural network that exploits the collected data for exploitation.
    \item Under mild assumptions of over-parameterized neural networks, we provide an instance-based $\widetilde{\mathcal{O}}(\sqrt{T})$  regret upper bound of for \sysn, where the complexity term in this bound is easier to interpret, and this bound is at least as good as the existing works.
    \item We conduct experiments on real-world datasets, showing that EE-Net outperforms baselines, including linear and neural versions of $\epsilon$-greedy, TS, and UCB.
\end{enumerate}
Next, we discuss the problem definition in Sec.\ref{sec:prob}, elaborate on the proposed \sysn in Sec.\ref{sec:method}, and present our theoretical analysis in Sec.\ref{sec:analysis}. In the end, we provide the empirical evaluation (Sec.\ref{sec:exp}) and conclusion (Sec.\ref{sec:conclusion}).

\section{Related Work}

Assuming linear reward realization, linear UCB-based bandit algorithms \citep{2011improved, 2016collaborative} and linear Thompson Sampling \citep{agrawal2013thompson, abeille2017linear} demonstrate impressive empirical performance across various real-world scenarios, achieving a near-optimal regret bound of \(\tilde{\mathcal{O}}(\sqrt{T})\). To relax the linearity assumption, \citet{filippi2010parametric} generalize the reward function to include both linear and non-linear components, adopting a UCB-based algorithm for estimation. Similarly, \citet{bubeck2011x} impose the Lipschitz property on the reward metric space and develop a hierarchical optimization approach for selection. \citet{valko2013finite} embed the reward function into a Reproducing Kernel Hilbert Space and propose kernelized TS/UCB bandit algorithms.

To learn non-linear reward functions, deep neural networks have been adapted to bandits in various ways. \citet{riquelme2018deep, lu2017ensemble} construct $L$-layer DNNs to learn arm embeddings and apply Thompson Sampling on the last layer for exploration. \citet{zhou2020neural} introduce a provable neural-based contextual bandit algorithm with a UCB exploration strategy, which \citet{zhang2020neural} extend to the TS framework. Their regret analysis leverages recent advances in the convergence theory of over-parameterized neural networks \citep{du2019gradient, allen2019convergence} and employ the Neural Tangent Kernel \citep{ntk2018neural, arora2019exact} to establish connections with linear contextual bandits \citep{2011improved}. \citet{ban2021convolutional} further integrate convolutional neural networks with UCB exploration for visual-aware applications. \citet{xu2020neural} utilize UCB-based exploration on the last layer of neural networks to reduce computational costs associated with gradient-based UCB. \citet{qi2022neural} explore the correlation among arms in contextual bandits. 
\citet{ban2021multi} introduce a multi-facet bandit problem where one bandit formulates one facet, respectively.
Unlike the aforementioned works, EE-Net retains the powerful representation capability of neural networks to learn the reward function and, for the first time, employs a separate neural network for exploration.

Recently, more ideas have been proposed by other neural bandit works \citep{ban2022eenet, kassraie2022neural,dai2022federated,gu2024batched,hwang2023combinatorial,neural_bandit_prompt_lin2023use}, and tailored for various application scenarios such as active learning \citep{neural_active_learning_bandit_wang2021neural,active_learning_improved_ban2022improved,ban2024neuralactive}, meta-learning\citep{ban2024meta,qi2024meta}, and bandit-based graph learning \citep{qi2022neural,GNB,GNN-PE_kassraie2022graph}.
\cite{ban2024pagerank} studies PageRank-based bandit approaches for link prediction in graphs.
\citet{online_regression_neural_bandit_deb2023contextual} use inverse reward gap of neural approximation for exploration, and \citet{neural_bandit_perturbed_reward_jia2021learning} achieve efficient exploration by adding perturbations to received rewards in the training process.
\cite{qi2024robust} studies neural contextual bandits in the presence of adversarial corruptions.
\citep{osband2023epistemic,osband2023approximate} propose a new method for approximating Thompson sampling using epistemic neural networks that are designed to produce accurate joint predictive distributions, which is further extended to the exploration in language models \citep{dwaracherla2024efficient}.
 \cite{yang2026your} identifies inherent bias in group-relative advantage estimators.
 \cite{lu2026contextual} studies contextual rollout bandits in reinforcement learning with verifiable rewards.
 \cite{huang2025adaptive} studies bandit-based adaptive sample scheduling for direct preference optimization.
 \cite{qi2026bilevel} studies bi-level hierarchical neural contextual bandits for online recommendation.
To enhancing the exploration in reinforcement learning, \citep{burda2018exploration}
involves using two neural networks: a fixed, randomly initialized target network and a predictor network trained to mimic the target's outputs. Then, the learnr explore the states where the prediction error between these networks is high, whereas EE-net integrates both exploitation and exploration directly into its neural network model. \citep{osband2018randomized,dwaracherla2022ensembles} propose using ensemble methods augmented with fixed prior functions or bootstrapping to quantify uncertainty and drive exploration in deep reinforcement learning, which differ from the exploitation–exploration neural networks in EE-Net.

For more variants of exploration strategies, GIOR \citep{kveton2019garbage} explores by randomizing the history of rewards with pseudo-rewards, ensuring optimism in its bootstrap mean estimates. Bayes-UCB \citep{kaufmann2012bayesian}, derived from the Bayesian index perspective, relies on posterior quantiles for exploration and has been shown to achieve asymptotic optimality. Top-Two sampling \citep{jourdan2022top,russo2016simple}, as an adaptation of Thompson Sampling, randomizes between a leader and a challenger arm, aiming to balance exploration and exploitation for best-arm identification.

While these methods employ different mechanisms to encourage optimism or randomized exploration, they generally operate in a single or random “direction” of exploration.  In contrast, our proposed EE-Net introduces a novel neural-based bi-directional exploration strategy, where the dual-network design allows EE-Net to capture exploration in both directions.

\section{Problem Definition} \label{sec:prob}

 Under the setting of stochastic contextual bandits, in $t$-th round, $t \in [T]$ where the sequence $[T] = [1, 2, \dots, T]$, the learner is presented with $n$ arms represented by $n$ context vectors, $\mathbf{X}_t = \{\bx_{t,1}, \dots, \bx_{t, n}\}$, $\bx_{t,i} \in \bbr^{d} $, $ \forall i \in [n]$.
 Then, the learner is compelled to select one arm $\bx_{t, \hi}, \hi \in [n]$, and observe the corresponding reward $r_{t,\hi}$.
 For each arm $\bx_{t,i}, i \in [n]$, its reward $r_{t,i}$ is assumed to be governed by the function: 
 \begin{equation} \label{eq:h}
     r_{t,i} = h(\bx_{t,i}) + \eta_{t,i},
 \end{equation}
where the \emph{unknown} reward function $h(\bx_{t,i})$ can be either linear or non-linear and $\eta_{t,i}$ is the noise term. Here, we assume the noise $\bbe[\eta_{t,i}]=0$, while related works such as \citep{zhou2020neural, ban2021multi, zhang2020neural} assume $\eta_{t,i}$ is conditioned zero-mean sub-Gaussian noise.
Finally, the pseudo \emph{regret} of $T$ rounds is defined as:
\begin{equation} 
\mathbf{R}_T = \mathbb{E} \left[ \sum_{t=1}^T (r_{t, \iast} - r_{t, \hi})   \right],
\end{equation}
where $\iast = \arg \max_{i \in [n]} h(\bx_{t,i})$ is the index of an arm with the maximal expected reward in round $t$. The goal of this problem is to minimize $\mathbf{R}_T$ by optimizing the selection policy.
 
 \para{Notation.} We denote by $\{\bx_\tau\}_{\tau=1}^t$ the sequence $(\bx_1, \dots, \bx_t)$. We use $\|v\|_2$ to denote the Euclidean norm for a vector $v$, and $\|\bw\|_2$ and $\| \bw  \|_F $ to denote the spectral and Frobenius norm for a matrix $\bw$. We use $\langle \cdot, \cdot \rangle$ to denote the standard inner product between two vectors or two matrices. We may use $\triangledown_{\btheta^1_t}f_1(\bx_{t,i})$ or  $\triangledown_{\btheta^1_t}f_1$  to represent the gradient $\triangledown_{\btheta^1_t}f_1(\bx_{t,i}; \btheta^1_t)$ for brevity. We use $\{\bx_\tau, r_\tau \}_{\tau =1}^t$ to represent the collected data up to round $t$.

\section{Proposed Method: EE-Net} \label{sec:method}

\sysn is composed of two neural networks to exploit the collected data and make exploration respectively. 
For convenience, denote the first neural network by   $f_1 (\cdot; \btheta^1) $, named by "exploitation network", and denote the second neural network by  $f_2 (\cdot; \btheta^2)$, named by "exploration network". The exploration network $f_2$ is the primary novel component of our proposed system.
Table \ref{tab:2} lists the structure of \sysn.

\para{(1) Exploitation network.} 
The exploitation network \( f_1 \) is a neural network to map arms to rewards. In round \( t \), the network is represented as \( f_1(\cdot ; \btheta^1_{t-1}) \), where the superscript of \( \btheta_{t-1}^1 \) denotes the network's index, and the subscript indicates the round when \( f_1 \)'s parameters were last updated. For an arm \( \bx_{t, i} \) where \( i \in [n] \), \( f_1(\bx_{t, i} ; \btheta^1_{t-1}) \) calculates the "exploitation score" for \( \bx_{t, i} \) based on historical data. Following a specific criterion, after selecting arm \( \bx_{t, \hi} \), a reward \( r_{t, \hi} \) is obtained.
Therefore, we can conduct stochastic gradient descent (SGD) to update $\btheta^1_{t-1}$ based on $(\bx_{t, \hi}, r_{t, \hi})$ and denote the updated parameters by $\btheta^1_t$.

\begin{algorithm}[t]
\renewcommand{\algorithmicrequire}{\textbf{Input:}}
\renewcommand{\algorithmicensure}{\textbf{Output:}}
\caption{ \sysn }\label{alg:main}
\begin{algorithmic}[1] 
\Require $f_1, f_2$,  $T$ (number of rounds),  $\eta_1$ (learning rate for $f_1$),  $\eta_2$ (learning rate for $f_2$),  $\phi$ (normalization operator, Remark \ref{remark1})  
\State Initialize $\btheta^1_0, \btheta^2_0$;  $\calh_0 = \emptyset$  
\For{ $t = 1 , 2, \dots, T$}
\State Observe $n$ arms $\{\bx_{t,1}, \dots, \bx_{t, n}\}$
\For{each $i \in [n]$ }
\State Compute $f_1(\bx_{t,i}; \btheta^1_{t-1}), f_2( \phi(\bx_{t,i}); \btheta^2_{t-1})$
\EndFor
\State $\hi  = \arg \max_{i \in [n]} \left(f_1(\bx_{t,i}; \btheta^1_{t-1} ) + f_2( \phi(\bx_{t,i}); \btheta^2_{t-1}) \right)
$     \ \ \
\State Play $\bx_{t, \hi}$ and observe reward $r_{t, \hi}$
\State $\mathcal{L}_1 = \frac{1}{2}  \left (f_1(\bx_{t, \hi}; \btheta^1_{t-1}) - r_{t, \hi} \right)^2$
\State   $\btheta^1_t = \btheta^1_{t-1} -  \eta_1  \tri_{\btheta^1_{t-1}}  \call_1$

\State $\mathcal{L}_2 = \frac{1}{2}  \left( f_2( \phi(\bx_{t, \hi})); \btheta^2_{t-1}) - ( r_{t, \hi} - f_1(\bx_{t, \hi}; \btheta^1_{t-1}) )  \right)^2$
\State $\btheta^2_t = \btheta^2_{t-1} -  \eta_2  \tri_{\btheta^2_{t-1}}  \call_2$
\EndFor
\end{algorithmic}
\end{algorithm}

\para{2) Exploration network.}  
Our exploration strategy is inspired by existing UCB-based neural bandits \citep{zhou2020neural, ban2021multi}. 
Given an arm $\bx_{t,i}$,  with high probability,  the following UCB form holds:
\begin{equation}
|h(\bx_{t,i}) - f_1(\bx_{t,i}; \btheta^1_{t-1})| \leq \kappa ( \nabla_{\btheta^1_{t-1}}f_1(\bx_{t,i}; \btheta^1_{t-1})),
\end{equation}
where $h$ is defined in Eq. (\ref{eq:h}) and $\kappa$ is an upper confidence bound represented by a function with respect to the gradient $\nabla_{\btheta^1_{t-1}}f_1$ (see more details and discussions in Appendix \ref{sec:ucb}).
Then we have the following definition.

\begin{definition} [Potential Gain]
In round $t$, given an arm $\bx_{t,i}$, we define $ h(\bx_{t,i}) - f_1(\bx_{t,i}; \btheta^1_{t-1})$ as the "\emph{expected potential gain}" for $\bx_{t,i}$ and $r_{t,i} - f_1(\bx_{t,i}; \btheta^1_{t-1})$ as the "\emph{potential gain}" for $\bx_{t,i}$.
\end{definition}

Let $y_{t,i} = r_{t,i} - f_1(\bx_{t,i}; \btheta^1_{t-1})$. 
When $y_{t,i} > 0$, the arm $\bx_{t,i}$ has positive potential gain compared to the estimated reward $f_1(\bx_{t,i}; \btheta^1_{t-1})$. 
A large positive $y_{t,i}$ makes the arm more suitable for upward exploration while 
a large negative $y_{t,i}$ makes the arm more suitable for downward exploration. 
In contrast, $y_{t,i}$ with a small absolute value makes the arm unsuitable for exploration.
Recall that traditional approaches such as UCB intend to estimate such potential gain $y_{t,i}$ using standard tools, e.g., Markov inequality and Hoeffding bounds from large deviation bounds.

Instead of calculating a large-deviation based statistical bound for $y_{t,i}$, we use a neural network $f_2(\cdot; \btheta^2)$ to represent $g$, where the input is $\nabla_{\btheta^1_{t-1}}f_1(\bx_{t,i})$ and the ground truth is $r_{t,i} - f_1(\bx_{t,i}; \btheta^1_{t-1})$. Adopting gradient $\nabla_{\btheta^1_{t-1}}f_1(\bx_{t,i})$ as the input is due to the fact 
that it incorporates two aspects of information: the features of the arm and the discriminative information of $f_1$.  

Moreover, in the upper bound of NeuralUCB \citep{zhou2020neural} or the variance of NeuralTS \citep{zhang2020neural}, there is a recursive term $\mathbf{A}_{t-1} =\mathbf{I} + \sum_{\tau=1}^{t-1} \nabla_{\btheta^1_{\tau-1}}f_1(\bx_{\tau, \hi}) \nabla_{\btheta^1_{\tau-1}}f_1(\bx_{\tau, \hi})^\top$ which is a function of past gradients up to $(t-1)$ and incorporates relevant historical information. On the contrary, in EE-Net, the recursive term which depends on past gradients is $\btheta^2_{t-1}$ in the exploration network $f_2$ because we have conducted gradient descent for $\btheta^2_{t-1}$ based on $\{\nabla_{\btheta^1_{\tau-1}}f_1(\bx_{\tau, \hi}) \}_{\tau =1}^{t-1}$. Therefore, this form $\btheta^2_{t-1}$ is similar to $\mathbf{A}_{t-1}$ in neuralUCB/TS, but EE-net does not (need to) make a specific assumption about the functional form of past gradients, and it is also more memory-efficient.

To summarize, in round \( t \), we define \( f_2(\nabla_{\btheta^1_{t-1}}f_1(\bx_{t,i}); \btheta^2_{t-1}) \) as the "exploration score" for \( \bx_{t,i} \) to facilitate adaptive exploration. This score reflects the potential gain of \( \bx_{t,i} \) compared to our current exploitation score \( f_1(\bx_{t,i}; \btheta^1_{t-1}) \). After receiving the reward \( r_t \), we can update \( \btheta^2 \) using gradient descent based on the collected training samples \( \{ \nabla_{\btheta^1_{\tau-1}}f_1(\bx_{\tau, \hi}), r_\tau - f_1(\bx_{\tau, \hi}; \btheta^1_{\tau-1}) \}_{\tau=1}^t \). Additionally, we propose two heuristic forms for \( f_2 \)'s ground-truth label: \( | r_{t,i} - f_1(\bx_{t,i}; \btheta^1_{t-1})| \) and \( \text{ReLU}(r_{t,i} - f_1(\bx_{t,i}; \btheta^1_{t-1})) \). Algorithm \ref{alg:main} depicts the workflow of this \sysn.



\begin{remark} [\textbf{Network structure}] \label{remark1}
The structures of the networks \( f_1 \) and \( f_2 \) can vary depending on the application. For instance, in vision tasks, \( f_1 \) can be implemented as transformers \citep{vaswani2017attention}. For the exploration network \( f_2 \), the input \( \nabla_{\btheta^1} f_1 \) might have high dimensions when the exploitation network \( f_1 \) is wide and deep, leading to substantial computational costs for \( f_2 \). 
To mitigate this issue, dimensionality reduction techniques can be used to obtain low-dimensional vectors of \( \nabla_{\btheta^1} f_1 \). In our experiments, we employed the locally linear embedding (LLE) method from \citep{roweis2000nonlinear} to reduce \( \nabla_{\btheta^1} f_1 \) to a 10-dimensional vector, which achieved the best performance among all baselines. We chose LLE for its ability to preserve local non-linear structures in the high-dimensional gradient space. The resulting embedding vector is denoted by \( \phi(\bx_{t, i}) \) after normalization.
\end{remark}

\begin{remark} [\textbf{Exploration direction}]
\sysn has the capability to determine the direction of exploration. Given an arm \( \bx_{t,i} \), if the estimate \( f_1(\bx_{t,i}) \) is \emph{lower} than the expected reward \( h(\bx_{t,i}) \), the learner should increase the likelihood of exploring \( \bx_{t,i} \) ("upward" exploration). Conversely, if \( f_1(\bx_{t,i}) \) is \emph{higher} than \( h(\bx_{t,i}) \), the learner should decrease the likelihood of exploring \( \bx_{t,i} \) ("downward" exploration). EE-Net employs the neural network \( f_2 \) to predict \( h(\bx_{t,i}) - f_1(\bx_{t,i}) \), which yields positive or negative scores and thus determines the exploration direction.
\end{remark}

\begin{remark} [\textbf{Space complexity}]
NeuralUCB and NeuralTS need to maintain the gradient outer product matrix (e.g., \( \mathbf{A}_t = \sum_{\tau=1}^t \nabla_{\btheta^1} f_1(\bx_{\tau, \hi}; \btheta^1_\tau) \nabla_{\btheta^1} f_1(\bx_{\tau, \hi}; \btheta^1_\tau)^\top \in \mathbb{R}^{p_1 \times p_1} \) and \( \btheta^1 \in \mathbb{R}^{p_1} \)), which has a space complexity of \( O(p_1^2) \) to store the outer product. In contrast, EE-Net does not require this matrix and treats \( \nabla_{\btheta^1} f_1 \) only as the input to \( f_2 \). Consequently, EE-Net reduces the space complexity from \( \mathcal{O}(p_1^2) \) to \( \mathcal{O}(p_1) \).
\end{remark}

\begin{table}[t]
	\caption{ Structure of \sysn (Round $t$).}\label{tab:2}
	\centering
	\begin{tabularx}{\textwidth}{c|X|X}
		\toprule
		   Input   &  Network &  Label \\
		\toprule  
		  $\{\bx_{\tau, \hi} \}_{\tau=1}^t$ & $f_1(\cdot; \btheta^1)$ (Exploitation)  &  $\{r_\tau\}_{\tau=1}^t$  \\
		  \midrule  
		$ \{  \nabla_{\btheta_{\tau-1}^1} f_1(\bx_{\tau, \hi}; \btheta_{\tau-1}^1)  \}_{\tau=1}^t$  & $f_2(\cdot; \btheta^2) $ (Exploration)    & $ \left \{  \left( r_\tau - f_1(\bx_{\tau, \hi}; \btheta_{\tau-1}^1 )  \right) \right \}_{\tau=1}^t$ \\
		\bottomrule
	\end{tabularx}
	\vspace{-1em}
\end{table}
\section{Regret Analysis} \label{sec:analysis}

In this section, we provide the regret analysis of \sysn. 
Then, for the analysis, we have the following assumption, which is a standard  input assumption in neural bandits and  over-parameterized neural networks \citep{zhou2020neural, allen2019convergence}. 

\begin{assumption}  \label{assum:2}
For any $t \in [T], i \in[n], \|\bx_{t, i}\|_2 = 1$,  and $r_{t,i} \in [0, 1]$. 
\end{assumption}

The analysis will focus on over-parameterized neural networks \citep{ntk2018neural,du2019gradient,allen2019convergence}. Given an input $\bx \in \bbr^d$, we define the fully-connected network $f$ with depth $L \geq 2$ and width $m$:
\begin{equation} \label{eq:structure}
f(\bx; \btheta) =  \bw_L \sigma ( \bw_{L-1}  \sigma (\bw_{L-2} \dots  \sigma(\bw_1 \bx) ))
\end{equation}
where $\sigma$ is the ReLU activation function,  $\bw_1 \in \bbr^{m \times d}$, $ \bw_l \in \bbr^{m \times m}$, for $2 \leq l \leq L-1$, $\bw^L \in \bbr^{1 \times m}$, and  $\btheta = [ \text{vec}(\bw_1)^\intercal,  \text{vec}(\bw_2)^\intercal, \dots, \text{vec}(\bw_L )^\intercal ]^\intercal \in \bbr^{p}$ .

\emph{Initialization}. For any $l \in [L-1]$, each entry of $\bw_l$ is drawn from the normal distribution $\mathcal{N}(0, \frac{2}{m})$ and $\bw_L$ is drawn from the normal distribution $\mathcal{N}(0, \frac{1}{m})$.
$f_1$ and $f_2$ both follow above network structure but with different input and output, denoted by $f_1(\bx; \btheta^1), \btheta^1 \in \bbr^{p_1}, f_2(\phi(\bx); \btheta^2), \btheta^2 \in \bbr^{p_2}$.
Recall that $\eta_1, \eta_2$ are the learning rates for $f_1, f_2$.

Before providing the following theorem, first, we present the definition of function class following \citep{cao2019generalization}, which is closely related to our regret analysis.
Given the parameter space of exploration network $f_2$, we define the following function class around initialization:
\[\calb(\btheta_0^2, \omega) = \{ \btheta \in \bbr^{p_2}:    \| \btheta - \btheta_0^2 \|_2 \leq \omega/ m^{1/4} \}. 
\]
We slightly abuse the notations. Let $(\bx_1, r_1),  (\bx_2, r_2), \dots, (\bx_{Tn}, r_{Tn})$ represent all the data in $T$ rounds.
Then, we have the following instance-dependent complexity term:
\begin{equation} \label{complexityfunction}
\Psi(\btheta^2_0, w) = \underset{\btheta \in  \calb(\btheta_0^2, \omega)}{\inf} \sum_{t=1}^{Tn} (f_2(\bx_t; \btheta) - r_t)^2 
\end{equation}

Then, we provide the following regret bound.
\begin{restatable}{theorem}{theomain}  \label{theo1}
For any $\delta \in (0, 1), R >0$, suppose $m  \geq  \Omega \left(  \text{poly} (T, L, R, n, \log (1/\delta) ) \right)$, $ \eta_1 =  \eta_2  =\frac{ R^2 }{\sqrt{m} }$. 
Then, with probability at least $1 - \delta$ over the initialization,  the pseudo regret of Algorithm \ref{alg:main} in $T$ rounds satisfies
\begin{equation} 
\mathbf{R}_T \leq  \sqrt{T} \cdot \calo \left( \sqrt{ \Psi(\btheta^2_0, R)} +  \sqrt{ 2 \log ( 1/\delta) }    \right) + \calo(1).
 \end{equation}
\end{restatable}

Under the similar assumptions in over-parameterized neural networks, the regret bounds of NeuralUCB \citep{zhou2020neural} and NeuralTS \citep{zhang2020neural} are both
\begin{equation} \label{eq:baselinebound}
\begin{aligned}
\mathbf{R}_T \leq \mathcal{O} \left( \sqrt{ \widetilde{d} T \log T  + S^2 } \right) \cdot \mathcal{O} \left( \sqrt{\widetilde{d} \log T} \right), \\
S = \sqrt{2 \mathbf{h}\mathbf{H}^{-1}\mathbf{h}}
\  \ \text{and} \ \  \widetilde{d} = \frac{\log \text{det}(\mathbf{I} + \mathbf{H}/\lambda)}{ \log( 1 + Tn/\lambda)} 
\end{aligned}
\end{equation}
where $ \widetilde{d}$ is the effective dimension,
$\mathbf{H}$ is the neural tangent kernel matrix (NTK, Def.\ref{def:NTK}) \citep{ntk2018neural, arora2019exact} formed by the arm contexts of $T$ rounds defined in \citep{zhou2020neural}, 
and $\lambda$ is a regularization parameter.
For the complexity term $S$, $\mathbf{h} = \{h(\bx_{i})\}_{i = 1}^{Tn}$ and it represent the complexity of data of $T$ rounds. Similarly, in linear contextual bandits, \citet{2011improved} achieves $\mathcal{O}( d \sqrt{T}\log T)$ and \citet{li2017provably} achieves $\mathcal{O}( \sqrt{dT}\log T)$.

The complexity term $\Psi(\btheta^2_0, R)$ in Theorem \ref{theo1} is easier to interpret.
The complexity term $S$ depends on the smallest eigenvalue of $\mathbf{H}$ and $\mathbf{h}$ and thus it is not straightforward to explain the physical meaning of $S$. Moreover, as $h(\cdot) \in [0, 1]$,  in the worst case,  $\|\mathbf{h}\|_2  = \Theta(T)$. Therefore, $S^2$ can in general grow linearly with $T$. In contrast, $\Psi(\btheta^2_0, R)$ represents the smallest squared regression error a function class achieves, as defined in Eq. \ref{complexityfunction}.
The parameter $R$ in Theorem \ref{theo1} shows the size of the function class we can use to achieve small $\Psi(\btheta^2_0, R)$. 
Theorem \ref{theo1} shows that \sysn can achieve the optimal $\widetilde{\calo}(\sqrt{T})$ regret upper bound, if the data can be "well-classified" by the over-parameterized neural network functions, i,e., $\Psi(\btheta^2_0, R)$ is a small constant.
The instance-dependent terms $\Psi(\btheta^2_0, R)$ controlled by $R$ also appears in works \citep{chen2021provable, cao2019generalization}, but their regret bounds directly depend on $R$, i.e., $\widetilde{\calo}(\sqrt{T}) + \calo(R)$ . This indicates that their regret bounds are invalid (change to $\calo(T)$) when $R$ has  $\calo(T)$. On the contrary, we remove the dependence of $R$ in Theorem \ref{theo1}, which enables us to choose $R$ with $\calo(T)$ to have broader function classes. This is more realistic, because the parameter space is usually much larger than the number of data points for neural networks \citep{deng2009imagenet}. \citep{foster2020beyond,foster2021efficient} provide an instance-dependent regret upper bound which requires an online regression oracle, but their analysis is finished in the parametric setting regarding the reward function.

\begin{remark} \textbf{Context assumption}.
Theorem \ref{theo1} is notable for not making any assumptions about the contexts \(\{\bx_t\}_{t=1}^{Tn}\) used in the problem. This lack of restriction allow the arms to be chosen repeatedly. In contrast, existing neural bandit algorithms, such as those in \citep{zhou2020neural, zhang2020neural, kassraie2022neural}, rely on Assumption \ref{assum:ntk} for the contexts. This assumption requires the Neural Tangent Kernel (NTK) Gram matrix formed by \(\{\bx_t\}_{t=1}^{Tn}\) to be positive-definite, which means no arm context can be repetitively observed. Consequently, the regret upper bounds of these algorithms can be easily disrupted by straightforward context attacks, such as creating two identical contexts with different rewards.
\end{remark}

\begin{remark} \textbf{Proof workflow}.
Compared to NeuralUCB/TS, our proof is directly built on recent advances in convergence theory \citep{allen2019convergence} and generalization bound \citep{cao2019generalization} of over-parameterized neural networks. Instead, the analysis for NeuralUCB/TS contains three parts of approximation error by calculating the distances among the expected reward, ridge regression, NTK function, and the network function. 
\end{remark}

The proof of Theorem \ref{theo1} is in Appendix \ref{sec:the1p} and mainly based on the following generalization bound for the exploration neural network. The bound results from an online-to-batch conversion with an instance-dependent complexity term controlled by deep neural networks.

\begin{restatable}{lemma}{newthetabound}
\label{lemma:newthetabound}
Suppose $m, \eta_1, \eta_2$ satisfies the conditions in Theorem \ref{theo1}.
In round $t \in [T]$, let
\[\hi = \arg \max_{i \in [k]} \left(   f_1(\bx_{t,\hi}; \btheta^1_{t-1}) + f_2(\phi(\bx_{t,\hi}); \btheta^2_{t-1})   \right),\]
and denote the policy by $\pi_t$.
Then, for any $\delta \in (0, 1), R >0 $,  with probability at least $1-\delta$, for $t \in [T]$, it holds uniformly
\begin{equation}
\begin{aligned}
\frac{1}{t} \sum_{\tau=1}^t \underset{ r_{\tau, \hi}}{\bbe} \left[   \left| f_1(\bx_{\tau, \hi}; \btheta^1_{\tau-1}) + f_2(\phi(\bx_{\tau, \hi}); \btheta^2_{\tau-1})   - r_{\tau, \hi} \right| \mid \pi_t, \calh_{\tau - 1} \right] \\
\leq \frac{ \sqrt{ \Psi(\btheta^2_0, R) } + \calo(1) } {\sqrt{t}} +  \sqrt{ \frac{2  \log ( \calo(1)/\delta) }{t}}.
\end{aligned}
\end{equation}
where $  \calh_{t} = \{\bx_{\tau, \hi}, r_{\tau, \hi} \}_{\tau =1}^t$ represents of historical data selected by ${\pi_\tau}$ and expectation is taken over the reward.
\end{restatable}

Lemma \ref{lemma:newthetabound} establishes an instance-dependent generalization bound for exploration networks, exhibiting a complexity term of $\mathcal{O}(t^{-1/2})$. We achieve this by working in the regression rather than the classification setting and utilizing the almost convexity of square loss. Note that the bound in Lemma \ref{lemma:newthetabound} holds in the setting of bounded (possibly random) rewards $r \in [0,1]$ instead of a fixed function in the conventional classification setting.

\subsection{Connection to Neural Tangent Kernel}

In addition, we provide one method to connect $\Psi(\btheta_0, R)$ with NTK, or $\widetilde{d}$ and $S$, where  $\btheta_0$ represents the initialized parameter. 
The following assumption is widely used in neural contextual bandits \citep{zhou2020neural, zhang2020neural, kassraie2022neural}. It requires that NTK gram matrix formed by all arm contexts has to be positive-definite, which also implies that no arm context is allowed to be repetitively observed.

\begin{definition} [NTK \cite{ntk2018neural, wang2021neural}] Let $\mathcal{N}$ denote the normal distribution.
Given the data instances $\{\bx_t\}_{t=1}^{Tn}$, for all $i, j \in [Tn]$,  define 
\[
\begin{aligned}
&\mathbf{H}_{i,j}^0 = \Sigma^{0}_{i,j} =  \langle \bx_i, \bx_j\rangle,   \ \ 
\mathbf{A}^{l}_{i,j} =
\begin{pmatrix}
\Sigma^{l}_{i,i} & \Sigma^{l}_{i,j} \\
\Sigma^{l}_{j,i} &  \Sigma^{l}_{j,j} 
\end{pmatrix} \\
&   \Sigma^{l}_{i,j} = 2 \mathbb{E}_{a, b \sim  \mathcal{N}(\mathbf{0}, \mathbf{A}_{i,j}^{l-1})}[ \sigma(a) \sigma(b)], \\ & \mathbf{H}_{i,j}^l = 2 \mathbf{H}_{i,j}^{l-1} \mathbb{E}_{a, b \sim \mathcal{N}(\mathbf{0}, \mathbf{A}_{i,j}^{l-1})}[ \sigma'(a) \sigma'(b)]  + \Sigma^{l}_{i,j}.
\end{aligned}
\]
Then, the NTK matrix is defined as $ \mathbf{H} =  (\mathbf{H}^L + \Sigma^{L})/2$.
\end{definition} \label{def:NTK}

\begin{assumption} \label{assum:ntk}
There exists $\lambda_0 > 0$, such that $\mathbf{H} \succeq \lambda_0 \mathbf{I}$
\end{assumption}

With this assumption, we can further bound $\Psi(\btheta_0, R)$ in Theorem \ref{theo1} as the following lemma.

\begin{lemma} \label{lemma:upperboundofStk}
Suppose Assumption \ref{assum:ntk} and conditions in Theorem \ref{theo1} holds where $m \geq \widetilde{\Omega} ( \text{poly}(T, L) \cdot n \lambda_0^{-1} \log (1/\delta))$. With probability at least $1 - \delta$ over the initialization, there exists $\btheta'  \in  B(\btheta_0, \widetilde{\Omega}(T^{3/2}))$, such that
\[
\begin{aligned}
\bbe[\Psi(\btheta_0,  \widetilde{\Omega}(T^{3/2}))] \leq   \bbe [ \sum_{t=1}^{Tn} (f(\bx_t; \btheta') - r_t)^2   ] \leq  \wcalo \left(\sqrt{ \widetilde{d}} + S  \right)^2 \cdot \widetilde{d}.
\end{aligned}
\]
\end{lemma}

Lemma \ref{lemma:upperboundofStk} provides an upper bound for $\Psi(\btheta_0, \widetilde{\Omega}(T^{3/2}))$ by setting $R \geq  \widetilde{\Omega}(T^{3/2})$ for a neural network model $f$.
This connection implies that
$\bbe[\mathbf{R}_T] \leq \wcalo(\sqrt{\hd T} (S + \sqrt{\hd}))$ in  Theorem \ref{theo1}
This also implies that 
$\bbe[\mathbf{R}_T]$ in Theorem \ref{theo1} is at least as good as the upper bounds in \citep{zhou2020neural,wang2021neural}.

\section{Experiments} \label{sec:exp}

In this section, we evaluate \sysn on four real-world datasets comparing with strong state-of-the-art baselines. We first present the setup of experiments, then show regret comparison and report ablation study.  Codes are publicly available\footnote{https://github.com/banyikun/EE-Net-ICLR-2022}. 



\para{Baselines}. To comprehensively evaluate \sysn, we choose $4$ neural-based bandit algorithms, one linear  and one kernelized bandit algorithms.
\begin{enumerate}
    \item LinUCB \citep{2010contextual} explicitly assumes the reward is a linear function of arm vector and unknown user parameter and then applies the ridge regression and un upper confidence bound to determine selected arm.
    \item KernelUCB \citep{valko2013finite} adopts a predefined kernel matrix on the reward space combined with a UCB-based exploration strategy.

    \item NeuralNTK is a variant of KernelUCB specialized to the Neural Tangent Kernel.
    
    \item Neural-Epsilon adapts the epsilon-greedy exploration strategy on exploitation network $f_1$. With probability $1-\epsilon$, the arm is selected by $\bx_t = \arg \max_{i \in [n]} f_1(\bx_{t, i}; \btheta^1)$ and with probability $\epsilon$, the arm is chosen randomly.
    \item NeuralUCB \citep{zhou2020neural} uses the exploitation network $f_1$ to learn the reward function coming with an UCB-based exploration strategy.
    \item NeuralTS \citep{zhang2020neural} adopts the exploitation network $f_1$ to learn the reward function coming with an Thompson Sampling exploration strategy.
\end{enumerate}
Note that we do not report the results of LinTS and KernelTS in the experiments, because LinTS and KernelTS have been significantly outperformed by NeuralTS \citep{zhang2020neural}.

\para{MNIST dataset}. 
MNIST is a well-known image dataset \citep{lecun1998gradient} for the 10-class classification problem. 
Following the evaluation setting of existing works \citep{valko2013finite, zhou2020neural, zhang2020neural},  we transform this classification problem into bandit problem. Consider an image $\bx \in \bbr^{d}$, we aim to classify it from $10$ classes. First, in each round, the image $\bx$ is transformed into $10$ arms and presented to the learner, represented by $10$ vectors in sequence $\bx_1 = (\bx, \mathbf{0}, \dots, \mathbf{0}), \bx_2 = (\mathbf{0}, \bx, \dots, \mathbf{0}), \dots, \bx_{10} = (\mathbf{0}, \mathbf{0}, \dots, \mathbf{\bx}) \in \bbr^{10d} $. The reward is defined as $1$ if the index of selected arm matches the index of $\bx$'s ground-truth class; Otherwise, the reward is $0$. 

\para{Yelp and Movielens datasets}. 
Yelp\footnote{\url{https://www.yelp.com/dataset}} is a dataset released in the Yelp dataset challenge, which consists of 4.7 million rating entries for  $1.57 \times 10^5$ restaurants by $1.18$ million users. MovieLens\citep{harper2015movielens} is a dataset consisting of $25$ million ratings between $1.6 \times 10^5$ users and $6 \times 10^4$ movies. 
We build the rating matrix by choosing the top $2000$ users and top $10000$ restaurants(movies) and use singular-value decomposition (SVD) to extract a $10$-dimension feature vector for each user and restaurant(movie).
In these two datasets, the bandit algorithm is to choose the restaurants(movies) with bad ratings.
This is similar to the recommendation of good restaurants by catching the bad restaurants.
We generate the reward by using the restaurant(movie)'s gained stars scored by the users. In each rating record, if the user scores a restaurant(movie) less than 2 stars (5 stars totally), its reward is $ 1$; Otherwise, its reward is $ 0$. 
In each round,  we set $10$ arms as follows: we randomly choose one with reward $1$ and randomly pick the other $9$ restaurants(movies) with $0$ rewards; then, the representation of each arm is the concatenation of the corresponding user feature vector and restaurant(movie) feature vector.

\para{Disin dataset}. Disin\citep{ahmed2018detecting} is a fake news dataset on kaggle\footnote{https://www.kaggle.com/clmentbisaillon/fake-and-real-news-dataset} including 12600 fake news articles and 12600 truthful news articles, where each article is represented by the text. To transform the text into vectors, we use the approach \citep{DBLP:conf/sigir/FuH21} to represent each article by a 300-dimension vector. Similarly, we form a 10-arm pool in each round, where 9 real news and 1 fake news are randomly selected. If the fake news is selected, the reward is $1$; Otherwise, the reward is $0$.

\begin{table}[t]
	\vspace{-1em}
	\caption{ Cumulative regret of all methods in 10000 rounds.}\label{tab:allregret}
	\vspace{1em}
	\centering
 \small
	\begin{tabularx}{ 1.0  \textwidth}{c|c|c|c|c}
		\toprule
		     & MNIST &  Disin & Movielens & Yelp \\
		\toprule  
		    LinUCB &  7863.2 $\pm$32 & 2457.9 $\pm$ 11 & 2143.5  $\pm$ 35 & 5917.0 $\pm$ 34      \\
                KenelUCB & 7635.3 $\pm$ 28 & 8219.7 $\pm$ 21 & 1723.4  $\pm$ 3 & 4872.3 $\pm$ 11 \\
                NeuralNTK &  4692.3 $\pm$ 78  &   6736.7 $\pm$ 49 &   1879.4 $\pm$ 13 &  5253.2 $\pm$ 41     \\ 
                Neural-$\epsilon$ & 1126.8 $\pm$ 6   & 734.2 $\pm$ 31 & 1573.4  $\pm$ 26 &  5276.1 $\pm$ 27 \\
                NeuralUCB & 943.5  $\pm$ 8  & 641.7 $\pm$ 23 & 1654.0  $\pm$ 31 & 4593.1 $\pm$ 13 \\
                NeuralTS & 965.8 $\pm$ 87 & 523.2 $\pm$ 43 & 1583.1  $\pm$ 23 & 4676.6  $\pm$ 7 \\
		  \midrule  
            EE-Net &  \textbf{842.3}$\pm$72(\textbf{10.7\%$\uparrow$}) & \textbf{476.4}$\pm$23(\textbf{8.9\%$\uparrow$})& \textbf{1472.4}$\pm$5(\textbf{6.4\%$\uparrow$})& \textbf{4403.1}$\pm$13(\textbf{4.1\%$\uparrow$}) \\
		\bottomrule
	\end{tabularx}
	\vspace{-1em}
\end{table}

\para{Configurations}. For LinUCB, following \citep{2010contextual},  we do a grid search for the exploration constant  $\alpha$ over $(0.01, 0.1, 1)$ which is to tune the scale of UCB. 
For KernelUCB \citep{valko2013finite}, we use the radial basis function kernel and stop adding contexts after 1000 rounds, following \citep{valko2013finite,zhou2020neural}.
For the regularization parameter $\lambda$ and exploration parameter $\nu$ in KernelUCB, we do the grid search for $\lambda$ over $(0.1, 1, 10)$ and for $\nu$ over $(0.01, 0.1, 1)$. 
For NeuralUCB and NeuralTS, following setting of \citep{zhou2020neural, zhang2020neural}, we use the exploiation network $f_1$ and conduct the grid search for the exploration parameter $\nu$ over $(0.001, 0.01, 0.1,  1 )$ and for the regularization parameter $\lambda$ over $(0.01, 0.1, 1)$. For Neural-$\epsilon$, we use the same neural network $f_1$ and do the grid search for the exploration probability $\epsilon $ over $( 0.01, 0.1, 0.2)$.
For the neural bandits NeuralUCB/TS, following their setting, as they have expensive computation cost to store and compute the whole gradient matrix, we use a diagonal matrix to make approximation.
For all neural networks, we conduct the grid search for learning rate over $( 0.01, 0.001, 0.0005, 0.0001)$.
For all grid-searched parameters, we choose the best of them for the comparison and report the averaged results of $10$ runs for all methods.
To compare fairly, for all the neural-based methods, including \sysn, the exploitation network $f_1$ is built by a 2-layer fully-connected network with 100 width. For the exploration network $f_2$, we use a 2-layer fully-connected network with 100 width as well. In the training process, we update neural networks according to all past observed data, following \citep{zhou2020neural}.

\begin{figure}[t] 
    \centering
    \vspace{-1em}
     \begin{subfigure}[b]{0.99\textwidth}
    \includegraphics[width= 0.47\textwidth]{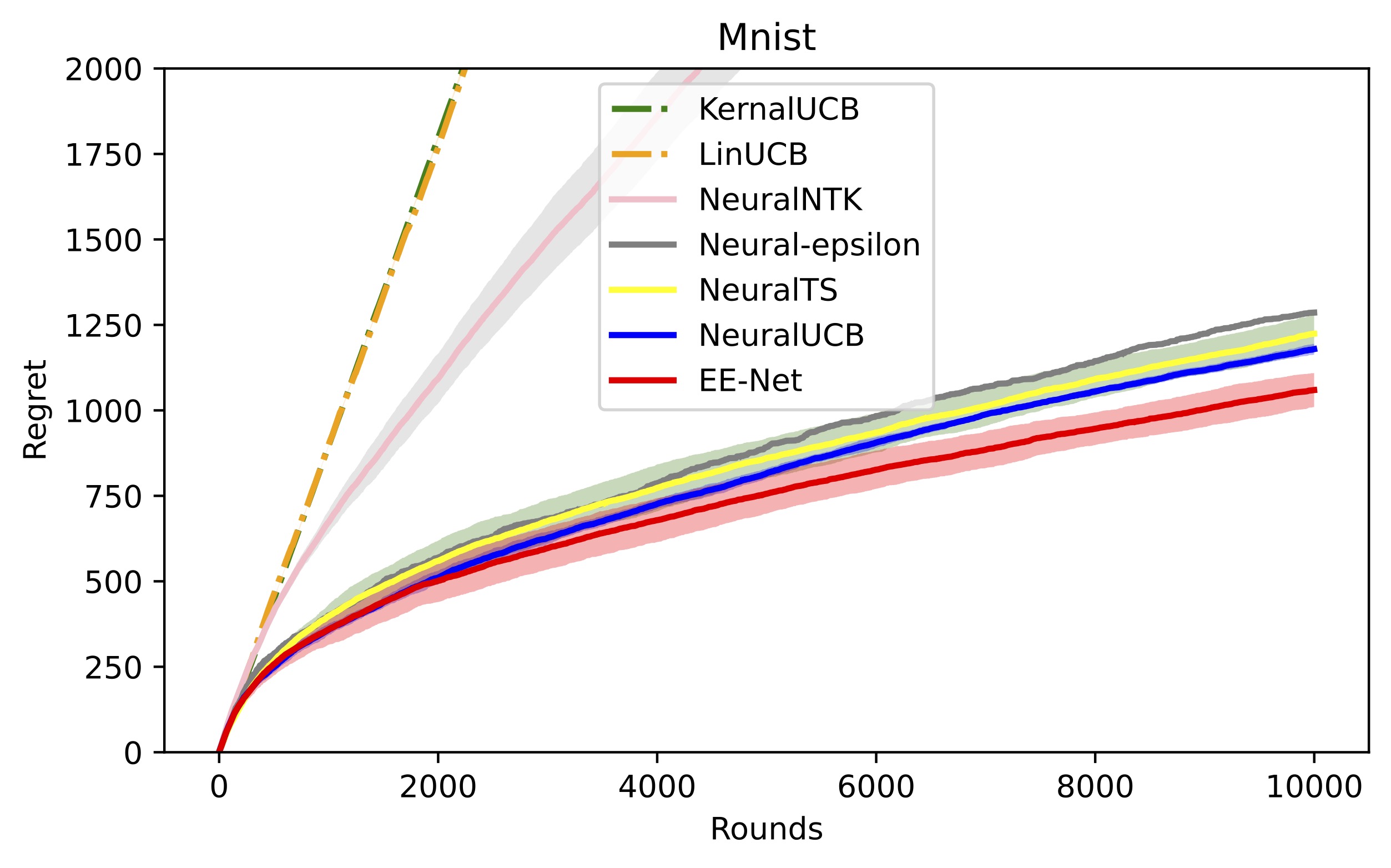}
     \includegraphics[width= 0.47\textwidth]{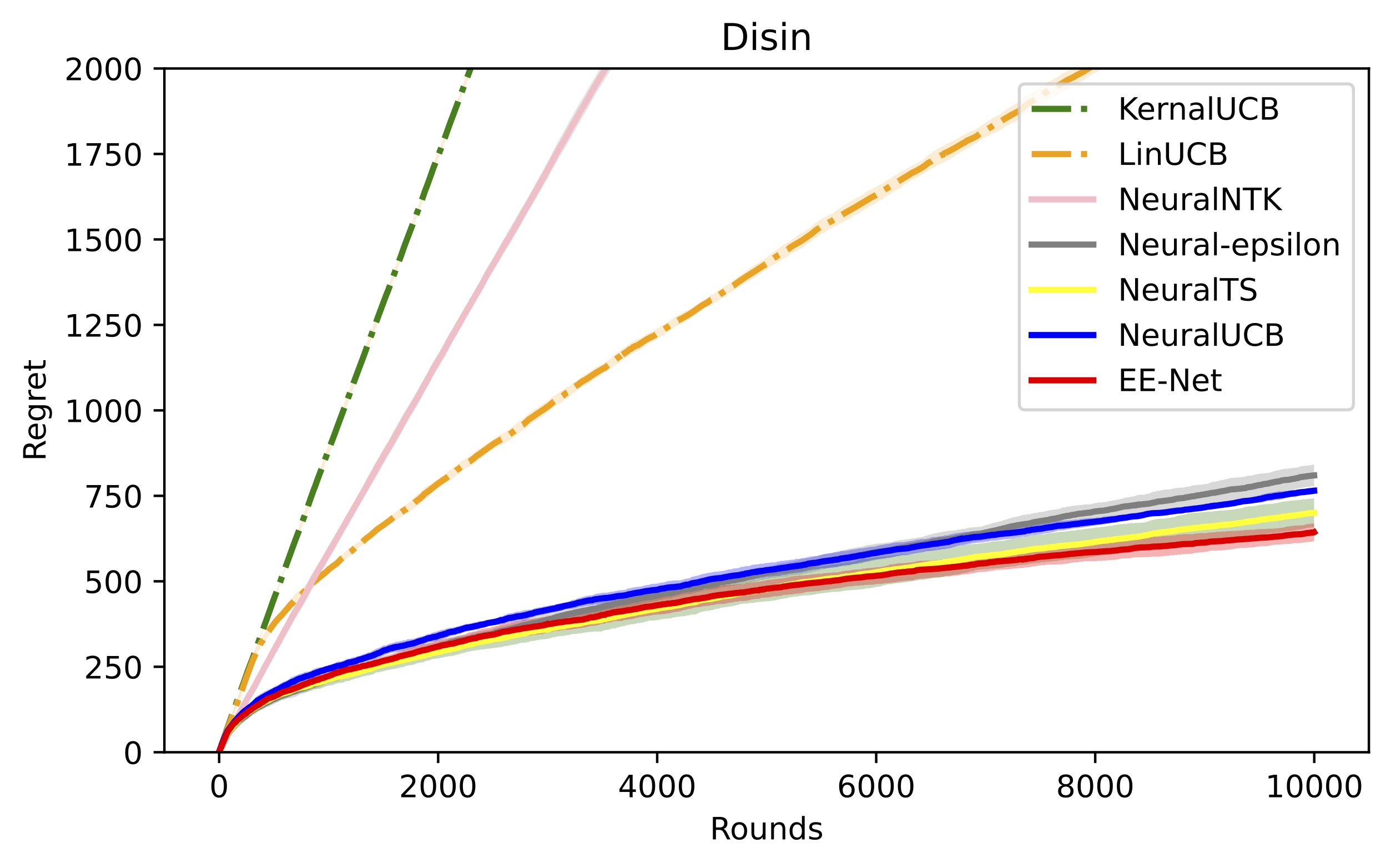}
    \centering
    \caption{ Regret comparison on MNIST and Disin.}
    \label{fig:1}
     \end{subfigure} 
      \begin{subfigure}[b]{0.99\textwidth}
       \includegraphics[width= 0.47\textwidth]{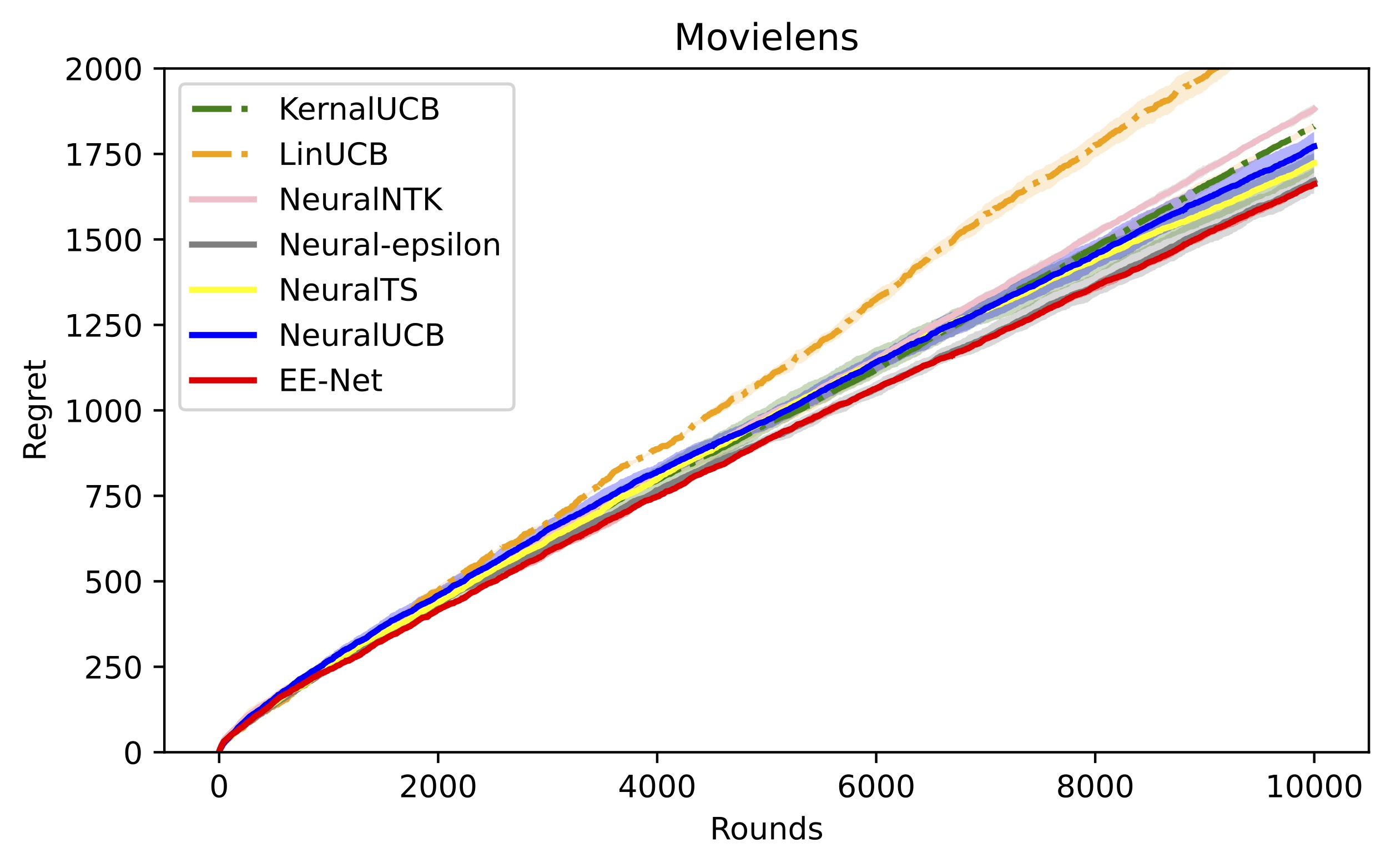}
     \includegraphics[width= 0.47\textwidth]{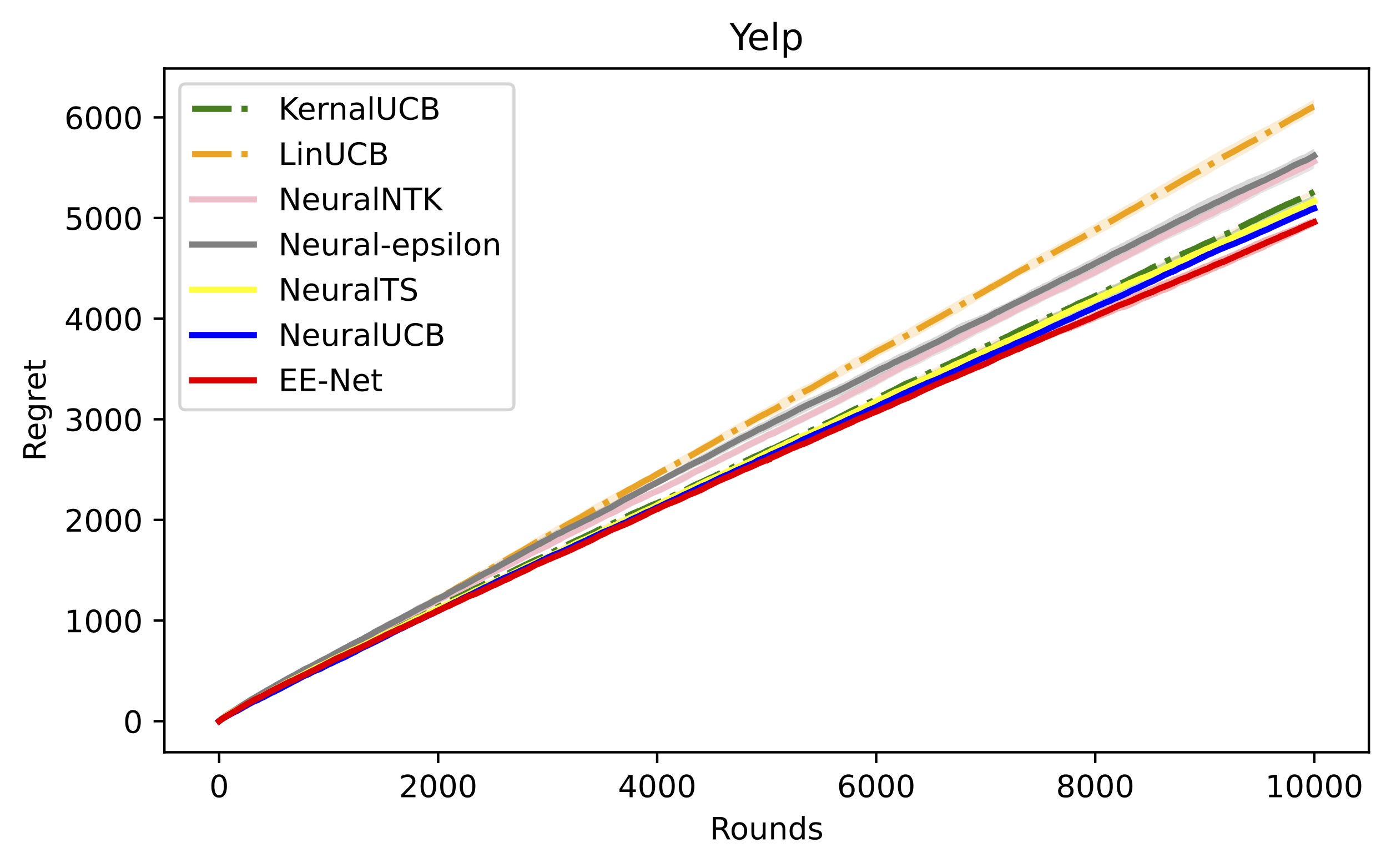}
    \centering
    \caption{Regret comparison on Movielens and Yelp.  }
      \label{fig:2}
      \end{subfigure}    
     \caption{ With the same exploitation network $f_1$, \sysn outperforms neural-based baselines. }
     \label{fig2_overall}
\end{figure}

\begin{figure}[h] 
     \includegraphics[width=0.47\columnwidth]{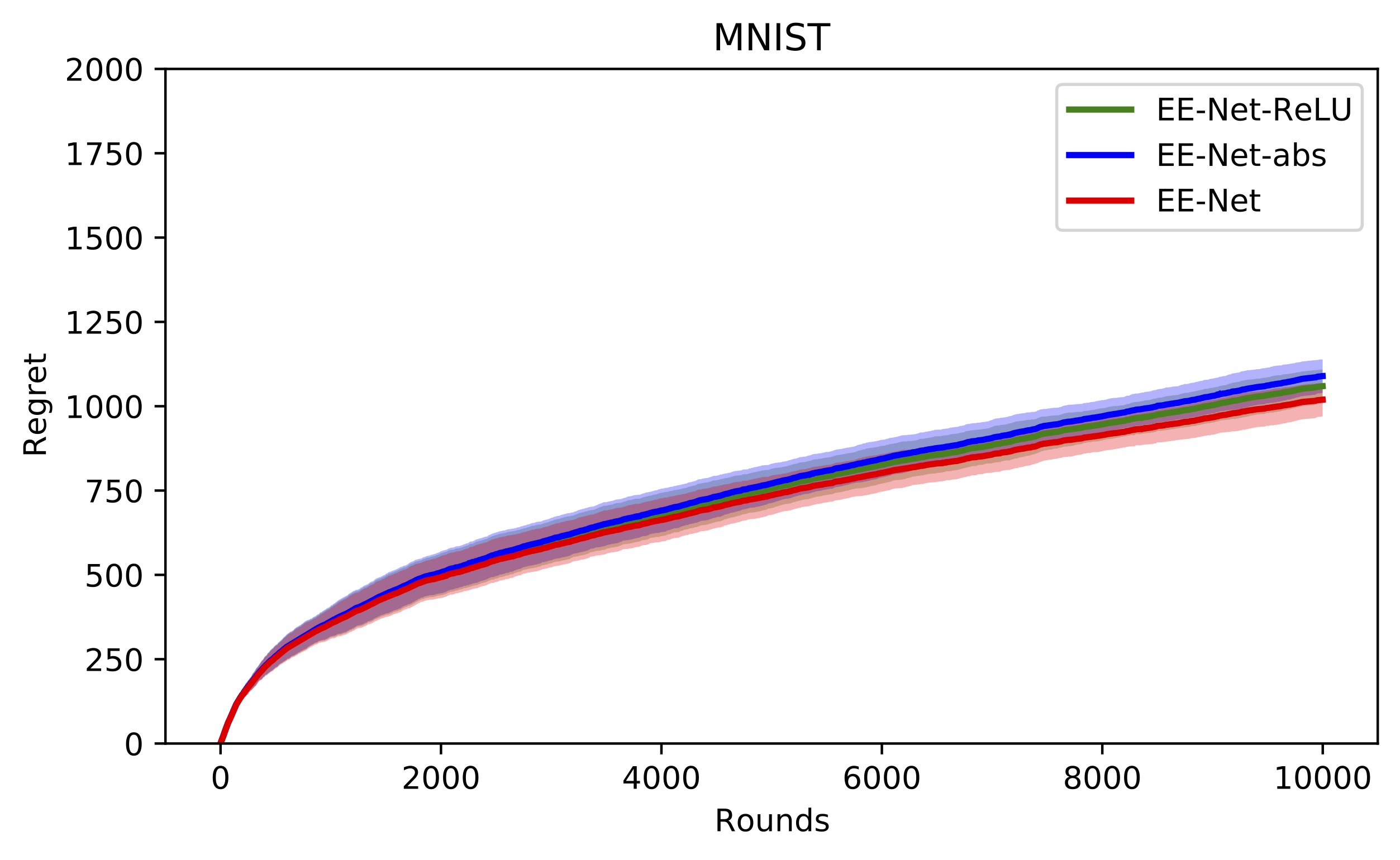}
    \includegraphics[width=0.47\columnwidth]{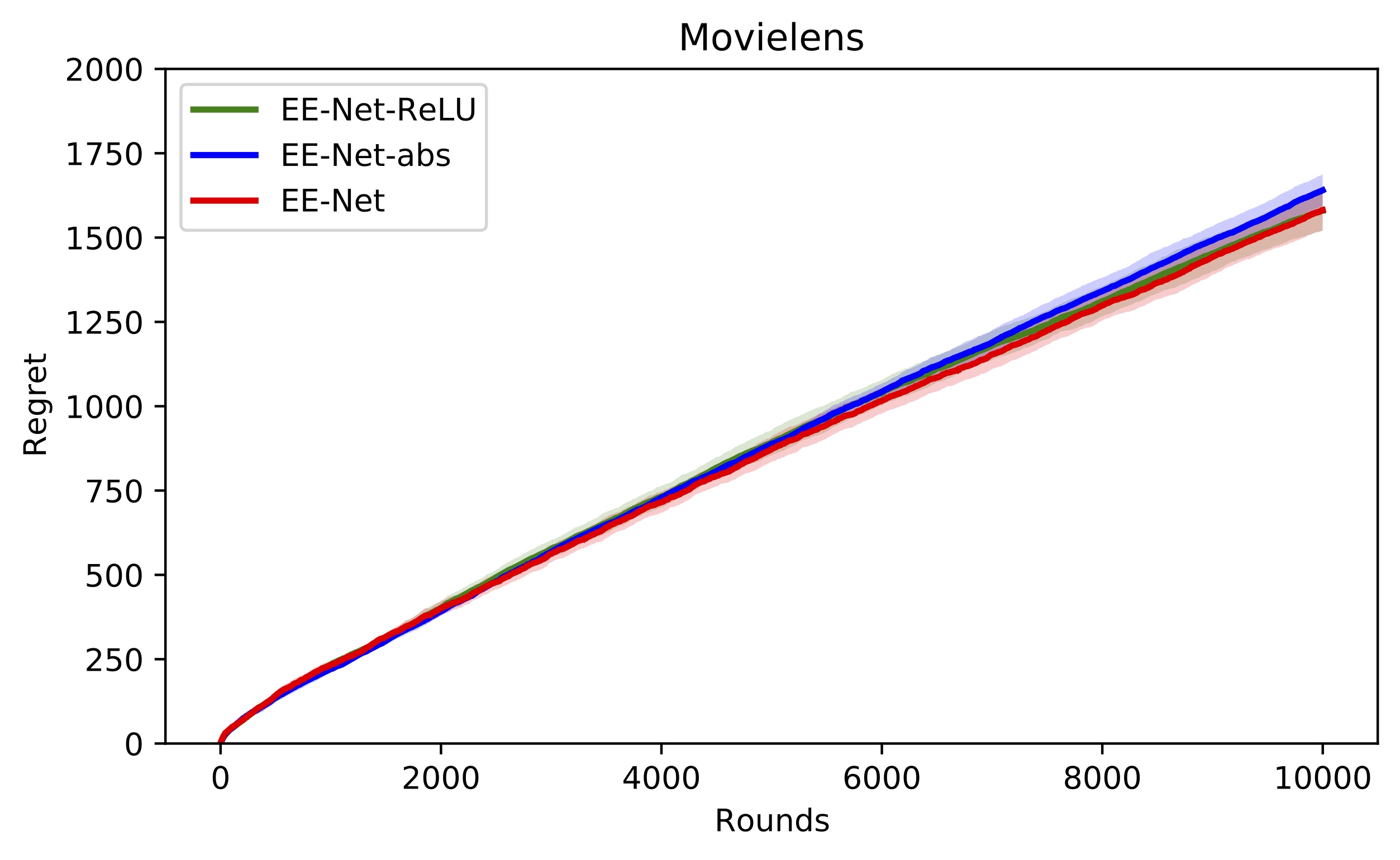}
    \centering
    \caption{ Ablation study on label function $y$ for $f_2$. \sysn denotes $y_1 = r - f_1$, EE-Net-abs denotes $y_2= | r  - f_1|$, and EE-Net-ReLU denotes $y_3 = \text{ReLU} (r- f_1)$. \sysn shows the best performance on these two datasets. }
       \label{fig:3}
\end{figure}

\para{Results}.
Figures \ref{fig2_overall} and \ref{fig:3} present the average cumulative regrets of all methods over 10,000 rounds, while Figures \ref{fig:1} and \ref{fig:2} compare regrets across four datasets. Table \ref{tab:allregret} reports the final regret of all methods. \sysn consistently outperforms all baselines.  
LinUCB and KernelUCB struggle due to their reliance on a simple linear reward function or predefined kernel, which fail to capture the complex reward structures in real-world datasets. This limitation is particularly evident in the MNIST and Disin datasets, where reward-arm correlations are neither linear nor simple mappings. As a result, these methods fail to effectively leverage past data and select optimal arms.  
Among neural-based bandit algorithms, NeuralNTK performs poorly in practice due to its reliance on over-parameterized assumptions, which are often impractical. Neural-$\epsilon$ relies on random exploration, failing to utilize available state information. NeuralUCB and NeuralTS attempt to address exploration through statistical confidence bounds. NeuralUCB computes a gradient-based upper confidence bound, while NeuralTS samples predicted rewards from a normal distribution with a gradient-based standard deviation. However, these approaches primarily account for worst-case deviations and may not adaptively estimate each arm’s potential gain.  
In contrast, \sysn leverages a neural network \( f_2 \) to learn arm potentials through powerful representation learning, enabling adaptive exploration. This allows \sysn to outperform state-of-the-art bandit algorithms. Additionally, while NeuralUCB and NeuralTS require tuning two parameters for confidence bounds or standard deviations across different tasks, \sysn simply sets up a neural network that autonomously learns optimal exploration strategies.


\para{Ablation study for $y$}. In this paper, we use $y_1 = r -  f_1$ to measure the potential gain of an arm, as the label of $f_2$. Moreover, we provide other two intuitive form  $y_2 = |r - f_1|$ and $y_3 = \text{ReLU}(r - f_1)$. 
Figure \ref{fig:3} shows the regret with different $y$, where "EE-Net" denotes our method with default $y_1$, "EE-Net-abs" represents the one with $y_2$ and "EE-Net-ReLU" is with $y_3$.  On Movielens and MNIST datasets, \sysn slightly outperforms EE-Net-abs and EE-Net-ReLU.  In fact, $y_1$ can effectively represent the positive potential gain and negative potential gain, such that $f_2$ intends to score the arm with positive gain higher and score the arm with negative gain lower. However, $y_2$ treats the positive/negative potential gain evenly, weakening the discriminative ability. $y_3$ can recognize the positive gain while neglecting the difference of negative gain. Therefore, $y_1$ usually is the most effective one for empirical performance.

\begin{table}[h]
	\caption{  Different dimensionality of input of exploration network }
	\centering
	\begin{tabular}{c|c|c|c|c}
		\toprule
		   Dimensionality   &  10 & 50 & 200& 500 \\ \hline
                Regret &   1472.3 $\pm$ 5   &   1463.5 $\pm$ 6.3   &  1452.1 $\pm$ 4.2    & 1467.6 $\pm$ 2.4  \\ 
                		\bottomrule
	\end{tabular}
        \label{tab:abondimension}
\end{table}

\para{Dimension reduction}. We conducted additional experiments with varying levels of dimension reduction for \(\nabla_{\btheta^1} f_1\). A grid search was performed over the set \(\{10, 20, 100, 500\}\) using the Movielens dataset. Table \ref{tab:abondimension} illustrates the performance changes with increasing dimensionality. The results indicate that higher dimensionality (from 10 to 200) can enhance performance because the feature embedding captures more information. However, this comes at the cost of increased computational requirements.
As the complexity of the exploitation network \( f_1 \) increases, a higher input dimensionality becomes necessary. However, a higher-dimensional input may demand a more complex architecture for \( f_2 \) to effectively learn the intricate mapping. This could explain the decline in performance observed when the dimensionality reaches 500.  
Therefore, a balance should be made between performance and computation complexity, depending on the specific models and applications.

\begin{figure}[h] 
    \includegraphics[width=0.67\columnwidth]{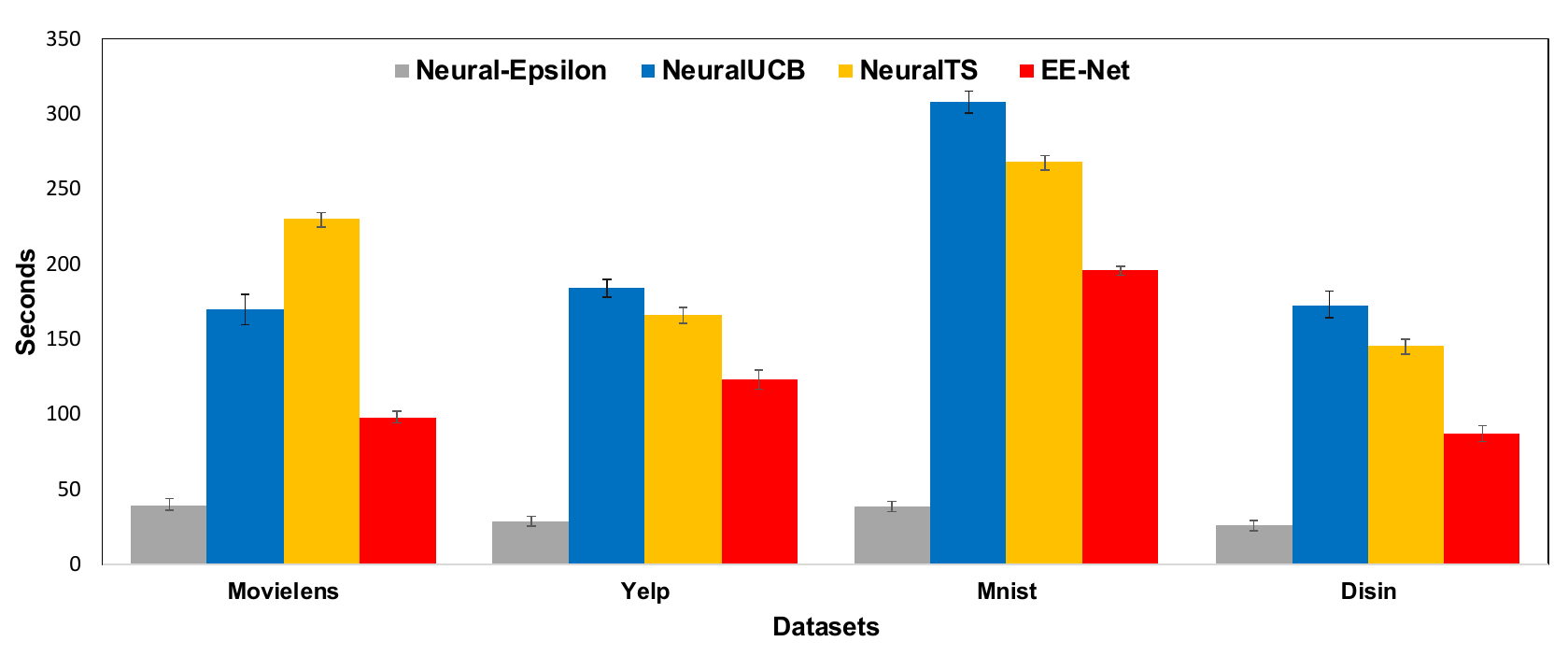}
    \centering
    \caption{Decision-making time}
       \label{fig:decisiontime}
\end{figure}

\para{Running time analysis}. 
During training, EE-Net incurs a higher cost since it requires training an additional neural network (the Exploration network), leading to approximately 38–46\% more computation compared to NeuralUCB and NeuralTS. In contrast, during inference, EE-Net is more efficient. NeuralUCB and NeuralTS must compute the inverse of the gradient outer product matrix at each decision step, whereas EE-Net only performs a single forward pass of the Exploration network. As a result, As shown in Figure \ref{fig:decisiontime}, EE-Net achieves 32–59\% faster inference compared to NeuralUCB and NeuralTS. This efficiency advantage is particularly important in real-world applications that demand fast decision-making and responsiveness.


\section{Conclusion} \label{sec:conclusion}

In this paper, we propose a novel exploration strategy, \sysn, by investigating the exploration direction in contextual bandits. In addition to a neural network that exploits collected data in past rounds, \sysn has another neural network to learn the potential gain compared to the current estimate for adaptive exploration. We provide an instance-dependent regret upper bound for \sysn and then use experiments to demonstrate its empirical performance.

\section*{Acknowledgements}
We are grateful to Shiliang Zuo and Yunzhe Qi for the valuable discussions in the revisions of manucript.
This research work is supported by National Science Foundation under Award No. IIS-1947203, IIS-2002540, IIS-2137468, IIS-1908104, OAC-1934634, and DBI2021898, and a grant from C3.ai. The views and conclusions are those of the authors and should not be interpreted as representing the official policies of the funding agencies or the government.

\bibliography{ref}

@inproceedings{2010contextual,
  title={A contextual-bandit approach to personalized news article recommendation},
  author={Li, Lihong and Chu, Wei and Langford, John and Schapire, Robert E},
  booktitle={Proceedings of the 19th international conference on World wide web},
  pages={661--670},
  year={2010}
}

@inproceedings{2011improved,
  title={Improved algorithms for linear stochastic bandits},
  author={Abbasi-Yadkori, Yasin and P{\'a}l, D{\'a}vid and Szepesv{\'a}ri, Csaba},
  booktitle={Advances in Neural Information Processing Systems},
  pages={2312--2320},
  year={2011}
}

@inproceedings{2016collaborative,
  title={Collaborative filtering bandits},
  author={Li, Shuai and Karatzoglou, Alexandros and Gentile, Claudio},
  booktitle={Proceedings of the 39th International ACM SIGIR conference on Research and Development in Information Retrieval},
  pages={539--548},
  year={2016}
}

@article{vaswani2017attention,
  title={Attention is all you need},
  author={Vaswani, Ashish and Shazeer, Noam and Parmar, Niki and Uszkoreit, Jakob and Jones, Llion and Gomez, Aidan N and Kaiser, {\L}ukasz and Polosukhin, Illia},
  journal={Advances in neural information processing systems},
  volume={30},
  year={2017}
}

@article{qi2024meta,
  title={Meta-learning with neural bandit scheduler},
  author={Qi, Yunzhe and Ban, Yikun and Wei, Tianxin and Zou, Jiaru and Yao, Huaxiu and He, Jingrui},
  journal={Advances in Neural Information Processing Systems},
  volume={36},
  year={2024}
}

@article{harper2015movielens,
  title={The movielens datasets: History and context},
  author={Harper, F Maxwell and Konstan, Joseph A},
  journal={Acm transactions on interactive intelligent systems (tiis)},
  volume={5},
  number={4},
  pages={1--19},
  year={2015},
  publisher={ACM New York, NY, USA}
}

@inproceedings{wu2016contextual,
  title={Contextual bandits in a collaborative environment},
  author={Wu, Qingyun and Wang, Huazheng and Gu, Quanquan and Wang, Hongning},
  booktitle={Proceedings of the 39th International ACM SIGIR conference on Research and Development in Information Retrieval},
  pages={529--538},
  year={2016}
}

@inproceedings{langford2008epoch,
  title={The epoch-greedy algorithm for multi-armed bandits with side information},
  author={Langford, John and Zhang, Tong},
  booktitle={Advances in neural information processing systems},
  pages={817--824},
  year={2008}
}

@inproceedings{filippi2010parametric,
  title={Parametric bandits: The generalized linear case},
  author={Filippi, Sarah and Cappe, Olivier and Garivier, Aur{\'e}lien and Szepesv{\'a}ri, Csaba},
  booktitle={Advances in Neural Information Processing Systems},
  pages={586--594},
  year={2010}
}

@inproceedings{chu2011contextual,
  title={Contextual bandits with linear payoff functions},
  author={Chu, Wei and Li, Lihong and Reyzin, Lev and Schapire, Robert},
  booktitle={Proceedings of the Fourteenth International Conference on Artificial Intelligence and Statistics},
  pages={208--214},
  year={2011}
}

@article{auer2002using,
  title={Using confidence bounds for exploitation-exploration trade-offs},
  author={Auer, Peter},
  journal={Journal of Machine Learning Research},
  volume={3},
  number={Nov},
  pages={397--422},
  year={2002}
}

@article{thompson1933likelihood,
  title={On the likelihood that one unknown probability exceeds another in view of the evidence of two samples},
  author={Thompson, William R},
  journal={Biometrika},
  volume={25},
  number={3/4},
  pages={285--294},
  year={1933},
  publisher={JSTOR}
}

@inproceedings{ban2020generic,
  title={Generic Outlier Detection in Multi-Armed Bandit},
  author={Ban, Yikun and He, Jingrui},
  booktitle={Proceedings of the 26th ACM SIGKDD International Conference on Knowledge Discovery \& Data Mining},
  pages={913--923},
  year={2020}
}

@article{valko2013finite,
  title={Finite-time analysis of kernelised contextual bandits},
  author={Valko, Michal and Korda, Nathaniel and Munos, R{\'e}mi and Flaounas, Ilias and Cristianini, Nelo},
  journal={arXiv preprint arXiv:1309.6869},
  year={2013}
}

@inproceedings{ntk2018neural,
  title={Neural tangent kernel: Convergence and generalization in neural networks},
  author={Jacot, Arthur and Gabriel, Franck and Hongler, Cl{\'e}ment},
  booktitle={Advances in neural information processing systems},
  pages={8571--8580},
  year={2018}
}

@inproceedings{arora2019exact,
  title={On exact computation with an infinitely wide neural net},
  author={Arora, Sanjeev and Du, Simon S and Hu, Wei and Li, Zhiyuan and Salakhutdinov, Russ R and Wang, Ruosong},
  booktitle={Advances in Neural Information Processing Systems},
  pages={8141--8150},
  year={2019}
}

@inproceedings{zhou2020neural,
  title={Neural contextual bandits with UCB-based exploration},
  author={Zhou, Dongruo and Li, Lihong and Gu, Quanquan},
  booktitle={International Conference on Machine Learning},
  pages={11492--11502},
  year={2020},
  organization={PMLR}
}

@article{bubeck2011x,
  title={X-Armed Bandits.},
  author={Bubeck, S{\'e}bastien and Munos, R{\'e}mi and Stoltz, Gilles and Szepesv{\'a}ri, Csaba},
  journal={Journal of Machine Learning Research},
  volume={12},
  number={5},
  year={2011}
}

@article{riquelme2018deep,
  title={Deep bayesian bandits showdown: An empirical comparison of bayesian deep networks for thompson sampling},
  author={Riquelme, Carlos and Tucker, George and Snoek, Jasper},
  journal={arXiv preprint arXiv:1802.09127},
  year={2018}
}

@inproceedings{allen2019convergence,
  title={A convergence theory for deep learning via over-parameterization},
  author={Allen-Zhu, Zeyuan and Li, Yuanzhi and Song, Zhao},
  booktitle={International Conference on Machine Learning},
  pages={242--252},
  year={2019},
  organization={PMLR}
}

@inproceedings{agrawal2013thompson,
  title={Thompson sampling for contextual bandits with linear payoffs},
  author={Agrawal, Shipra and Goyal, Navin},
  booktitle={International Conference on Machine Learning},
  pages={127--135},
  year={2013},
  organization={PMLR}
}

@article{roweis2000nonlinear,
  title={Nonlinear dimensionality reduction by locally linear embedding},
  author={Roweis, Sam T and Saul, Lawrence K},
  journal={science},
  volume={290},
  number={5500},
  pages={2323--2326},
  year={2000},
  publisher={American Association for the Advancement of Science}
}

@article{lecun1998gradient,
  title={Gradient-based learning applied to document recognition},
  author={LeCun, Yann and Bottou, L{\'e}on and Bengio, Yoshua and Haffner, Patrick},
  journal={Proceedings of the IEEE},
  volume={86},
  number={11},
  pages={2278--2324},
  year={1998},
  publisher={Ieee}
}

@inproceedings{du2019gradient,
  title={Gradient descent finds global minima of deep neural networks},
  author={Du, Simon and Lee, Jason and Li, Haochuan and Wang, Liwei and Zhai, Xiyu},
  booktitle={International Conference on Machine Learning},
  pages={1675--1685},
  year={2019},
  organization={PMLR}
}

@inproceedings{ban2021local,
  title={Local Clustering in Contextual Multi-Armed Bandits},
  author={Ban, Yikun and He, Jingrui},
  booktitle={Proceedings of the Web Conference 2021},
  pages={2335--2346},
  year={2021}
}

@inproceedings{zhang2020neural,
title={Neural Thompson Sampling},
author={Weitong Zhang and Dongruo Zhou and Lihong Li and Quanquan Gu},
booktitle={International Conference on Learning Representations},
year={2021},
}

@inproceedings{ban2021multi,
  title={Multi-facet Contextual Bandits: A Neural Network Perspective},
  author={Ban, Yikun and He, Jingrui and Cook, Curtiss B},
  booktitle = {The 27th {ACM} {SIGKDD} Conference on Knowledge Discovery and Data Mining, Virtual Event, Singapore, August 14-18, 2021},
  pages = {35--45},
  year      = {2021},
}

@article{cao2019generalization,
  title={Generalization bounds of stochastic gradient descent for wide and deep neural networks},
  author={Cao, Yuan and Gu, Quanquan},
  journal={Advances in Neural Information Processing Systems},
  volume={32},
  pages={10836--10846},
  year={2019}
}

@article{GNN-PE_kassraie2022graph,
  title={Graph Neural Network Bandits},
  author={Kassraie, Parnian and Krause, Andreas and Bogunovic, Ilija},
  journal={arXiv preprint arXiv:2207.06456},
  year={2022}
}

@article{online_regression_neural_bandit_deb2023contextual,
  title={Contextual bandits with online neural regression},
  author={Deb, Rohan and Ban, Yikun and Zuo, Shiliang and He, Jingrui and Banerjee, Arindam},
  journal={arXiv preprint arXiv:2312.07145},
  year={2023}
}

@inproceedings{ban2024neuralp,
  title={Neural Contextual Bandits for Personalized Recommendation},
  author={Ban, Yikun and Qi, Yunzhe and He, Jingrui},
  booktitle={Companion Proceedings of the ACM on Web Conference 2024},
  pages={1246--1249},
  year={2024}
}

@inproceedings{neural_bandit_perturbed_reward_jia2021learning,
  title={Learning Neural Contextual Bandits through Perturbed Rewards},
  author={Jia, Yiling and ZHANG, Weitong and Zhou, Dongruo and Gu, Quanquan and Wang, Hongning},
  booktitle={International Conference on Learning Representations},
  year={2021}
}

@inproceedings{GNB,
author = {Qi, Yunzhe and Ban, Yikun and He, Jingrui},
title = {Graph Neural Bandits},
year = {2023},
isbn = {9798400701030},
publisher = {Association for Computing Machinery},
address = {New York, NY, USA},
url = {https://doi.org/10.1145/3580305.3599371},
doi = {10.1145/3580305.3599371},
booktitle = {Proceedings of the 29th ACM SIGKDD Conference on Knowledge Discovery and Data Mining},
pages = {1920–1931},
numpages = {12},
location = {, Long Beach, CA, USA, },
series = {KDD '23}
}

@article{active_learning_improved_ban2022improved,
  title={Improved algorithms for neural active learning},
  author={Ban, Yikun and Zhang, Yuheng and Tong, Hanghang and Banerjee, Arindam and He, Jingrui},
  journal={Advances in Neural Information Processing Systems},
  volume={35},
  pages={27497--27509},
  year={2022}
}

@article{ban2024neuralactive,
  title={Neural active learning beyond bandits},
  author={Ban, Yikun and Agarwal, Ishika and Wu, Ziwei and Zhu, Yada and Weldemariam, Kommy and Tong, Hanghang and He, Jingrui},
  journal={arXiv preprint arXiv:2404.12522},
  year={2024}
}

@article{neural_active_learning_bandit_wang2021neural,
  title={Neural active learning with performance guarantees},
  author={Wang, Zhilei and Awasthi, Pranjal and Dann, Christoph and Sekhari, Ayush and Gentile, Claudio},
  journal={Advances in Neural Information Processing Systems},
  volume={34},
  pages={7510--7521},
  year={2021}
}

@article{neural_bandit_prompt_lin2023use,
  title={Use Your INSTINCT: INSTruction optimization usIng Neural bandits Coupled with Transformers},
  author={Lin, Xiaoqiang and Wu, Zhaoxuan and Dai, Zhongxiang and Hu, Wenyang and Shu, Yao and Ng, See-Kiong and Jaillet, Patrick and Low, Bryan Kian Hsiang},
  journal={arXiv preprint arXiv:2310.02905},
  year={2023}
}

@article{ahmed2018detecting,
  title={Detecting opinion spams and fake news using text classification},
  author={Ahmed, Hadeer and Traore, Issa and Saad, Sherif},
  journal={Security and Privacy},
  volume={1},
  number={1},
  pages={e9},
  year={2018},
  publisher={Wiley Online Library}
}

@inproceedings{DBLP:conf/sigir/FuH21,
  author    = {Dongqi Fu and
               Jingrui He},
  title     = {{SDG:} {A} Simplified and Dynamic Graph Neural Network},
  booktitle = {{SIGIR} '21: The 44th International {ACM} {SIGIR} Conference on Research
               and Development in Information Retrieval, Virtual Event, Canada, July
               11-15, 2021},
  pages     = {2273--2277},
  publisher = {{ACM}},
  year      = {2021},
}

@book{lattimore2020bandit,
  title={Bandit algorithms},
  author={Lattimore, Tor and Szepesv{\'a}ri, Csaba},
  year={2020},
  publisher={Cambridge University Press}
}

@inproceedings{abeille2017linear,
  title={Linear thompson sampling revisited},
  author={Abeille, Marc and Lazaric, Alessandro},
  booktitle={Artificial Intelligence and Statistics},
  pages={176--184},
  year={2017},
  organization={PMLR}
}

@article{lu2017ensemble,
  title={Ensemble sampling},
  author={Lu, Xiuyuan and Van Roy, Benjamin},
  journal={arXiv preprint arXiv:1705.07347},
  year={2017}
}

@article{ban2021convolutional,
  title={Convolutional Neural Bandit: Provable Algorithm for Visual-aware Advertising},
  author={Ban, Yikun and He, Jingrui},
  journal={arXiv preprint arXiv:2107.07438},
  year={2021}
}

@article{xu2020neural,
  title={Neural contextual bandits with deep representation and shallow exploration},
  author={Xu, Pan and Wen, Zheng and Zhao, Handong and Gu, Quanquan},
  journal={arXiv preprint arXiv:2012.01780},
  year={2020}
}

@inproceedings{li2017provably,
  title={Provably optimal algorithms for generalized linear contextual bandits},
  author={Li, Lihong and Lu, Yu and Zhou, Dengyong},
  booktitle={International Conference on Machine Learning},
  pages={2071--2080},
  year={2017},
  organization={PMLR}
}

@inproceedings{deng2009imagenet,
  title={Imagenet: A large-scale hierarchical image database},
  author={Deng, Jia and Dong, Wei and Socher, Richard and Li, Li-Jia and Li, Kai and Fei-Fei, Li},
  booktitle={2009 IEEE conference on computer vision and pattern recognition},
  pages={248--255},
  year={2009},
  organization={Ieee}
}

@article{chen2021provable,
  title={Provable regret bounds for deep online learning and control},
  author={Chen, Xinyi and Minasyan, Edgar and Lee, Jason D and Hazan, Elad},
  journal={arXiv preprint arXiv:2110.07807},
  year={2021}
}

@article{wang2021neural,
  title={Neural Active Learning with Performance Guarantees},
  author={Wang, Zhilei and Awasthi, Pranjal and Dann, Christoph and Sekhari, Ayush and Gentile, Claudio},
  journal={Advances in Neural Information Processing Systems},
  volume={34},
  year={2021}
}

@inproceedings{foster2020beyond,
  title={Beyond ucb: Optimal and efficient contextual bandits with regression oracles},
  author={Foster, Dylan and Rakhlin, Alexander},
  booktitle={International Conference on Machine Learning},
  pages={3199--3210},
  year={2020},
  organization={PMLR}
}

@inproceedings{bouneffouf2020survey,
  title={Survey on applications of multi-armed and contextual bandits},
  author={Bouneffouf, Djallel and Rish, Irina and Aggarwal, Charu},
  booktitle={2020 IEEE Congress on Evolutionary Computation (CEC)},
  pages={1--8},
  year={2020},
  organization={IEEE}
}

@article{slivkins2019introduction,
  title={Introduction to multi-armed bandits},
  author={Slivkins, Aleksandrs and others},
  journal={Foundations and Trends{\textregistered} in Machine Learning},
  volume={12},
  number={1-2},
  pages={1--286},
  year={2019},
  publisher={Now Publishers, Inc.}
}

@article{foster2021efficient,
  title={Efficient first-order contextual bandits: Prediction, allocation, and triangular discrimination},
  author={Foster, Dylan J and Krishnamurthy, Akshay},
  journal={Advances in Neural Information Processing Systems},
  volume={34},
  year={2021}
}

@inproceedings{
ban2022eenet,
title={{EE}-Net: Exploitation-Exploration Neural Networks in Contextual Bandits},
author={Yikun Ban and Yuchen Yan and Arindam Banerjee and Jingrui He},
booktitle={International Conference on Learning Representations},
year={2022},
url={https://openreview.net/forum?id=X_ch3VrNSRg}
}

@inproceedings{kveton2021meta,
  title={Meta-thompson sampling},
  author={Kveton, Branislav and Konobeev, Mikhail and Zaheer, Manzil and Hsu, Chih-wei and Mladenov, Martin and Boutilier, Craig and Szepesvari, Csaba},
  booktitle={International Conference on Machine Learning},
  pages={5884--5893},
  year={2021},
  organization={PMLR}
}

@inproceedings{kassraie2022neural,
  title={Neural contextual bandits without regret},
  author={Kassraie, Parnian and Krause, Andreas},
  booktitle={International Conference on Artificial Intelligence and Statistics},
  pages={240--278},
  year={2022},
  organization={PMLR}
}

@article{gu2024batched,
  title={Batched neural bandits},
  author={Gu, Quanquan and Karbasi, Amin and Khosravi, Khashayar and Mirrokni, Vahab and Zhou, Dongruo},
  journal={ACM/IMS Journal of Data Science},
  volume={1},
  number={1},
  pages={1--18},
  year={2024},
  publisher={ACM New York, NY}
}

@inproceedings{hwang2023combinatorial,
  title={Combinatorial neural bandits},
  author={Hwang, Taehyun and Chai, Kyuwook and Oh, Min-hwan},
  booktitle={International Conference on Machine Learning},
  pages={14203--14236},
  year={2023},
  organization={PMLR}
}

@article{dai2022federated,
  title={Federated neural bandit},
  author={Dai, Zhongxiang and Shu, Yao and Verma, Arun and Fan, Flint Xiaofeng and Low, Bryan Kian Hsiang and Jaillet, Patrick},
  journal={arXiv preprint arXiv:2205.14309},
  year={2022}
}

@inproceedings{qi2022neural,
  title={Neural Bandit with Arm Group Graph},
  author={Qi, Yunzhe and Ban, Yikun and He, Jingrui},
  booktitle={Proceedings of the 28th ACM SIGKDD Conference on Knowledge Discovery and Data Mining},
  pages={1379--1389},
  year={2022}
}

@inproceedings{osband2023approximate,
  title={Approximate thompson sampling via epistemic neural networks},
  author={Osband, Ian and Wen, Zheng and Asghari, Seyed Mohammad and Dwaracherla, Vikranth and Ibrahimi, Morteza and Lu, Xiuyuan and Van Roy, Benjamin},
  booktitle={Uncertainty in Artificial Intelligence},
  pages={1586--1595},
  year={2023},
  organization={PMLR}
}

@article{osband2023epistemic,
  title={Epistemic neural networks},
  author={Osband, Ian and Wen, Zheng and Asghari, Seyed Mohammad and Dwaracherla, Vikranth and Ibrahimi, Morteza and Lu, Xiuyuan and Van Roy, Benjamin},
  journal={Advances in Neural Information Processing Systems},
  volume={36},
  pages={2795--2823},
  year={2023}
}

@article{dwaracherla2024efficient,
  title={Efficient exploration for llms},
  author={Dwaracherla, Vikranth and Asghari, Seyed Mohammad and Hao, Botao and Van Roy, Benjamin},
  journal={arXiv preprint arXiv:2402.00396},
  year={2024}
}

@article{burda2018exploration,
  title={Exploration by random network distillation},
  author={Burda, Yuri and Edwards, Harrison and Storkey, Amos and Klimov, Oleg},
  journal={arXiv preprint arXiv:1810.12894},
  year={2018}
}

@article{osband2018randomized,
  title={Randomized prior functions for deep reinforcement learning},
  author={Osband, Ian and Aslanides, John and Cassirer, Albin},
  journal={Advances in Neural Information Processing Systems},
  volume={31},
  year={2018}
}

@article{dwaracherla2022ensembles,
  title={Ensembles for uncertainty estimation: Benefits of prior functions and bootstrapping},
  author={Dwaracherla, Vikranth and Wen, Zheng and Osband, Ian and Lu, Xiuyuan and Asghari, Seyed Mohammad and Van Roy, Benjamin},
  journal={arXiv preprint arXiv:2206.03633},
  year={2022}
}

@inproceedings{kveton2019garbage,
  title={Garbage in, reward out: Bootstrapping exploration in multi-armed bandits},
  author={Kveton, Branislav and Szepesvari, Csaba and Vaswani, Sharan and Wen, Zheng and Lattimore, Tor and Ghavamzadeh, Mohammad},
  booktitle={International Conference on Machine Learning},
  pages={3601--3610},
  year={2019},
  organization={PMLR}
}

@inproceedings{kaufmann2012bayesian,
  title={On Bayesian upper confidence bounds for bandit problems},
  author={Kaufmann, Emilie and Capp{\'e}, Olivier and Garivier, Aur{\'e}lien},
  booktitle={Artificial intelligence and statistics},
  pages={592--600},
  year={2012},
  organization={PMLR}
}

@article{jourdan2022top,
  title={Top two algorithms revisited},
  author={Jourdan, Marc and Degenne, R{\'e}my and Baudry, Dorian and de Heide, Rianne and Kaufmann, Emilie},
  journal={Advances in Neural Information Processing Systems},
  volume={35},
  pages={26791--26803},
  year={2022}
}

@inproceedings{russo2016simple,
  title={Simple bayesian algorithms for best arm identification},
  author={Russo, Daniel},
  booktitle={Conference on learning theory},
  pages={1417--1418},
  year={2016},
  organization={PMLR}
}

@inproceedings{ban2024meta,
  title={Meta clustering of neural bandits},
  author={Ban, Yikun and Qi, Yunzhe and Wei, Tianxin and Liu, Lihui and He, Jingrui},
  booktitle={Proceedings of the 30th ACM SIGKDD Conference on Knowledge Discovery and Data Mining},
  pages={95--106},
  year={2024}
}

@article{ban2024pagerank,
  title={Pagerank bandits for link prediction},
  author={Ban, Yikun and Zou, Jiaru and Li, Zihao and Qi, Yunzhe and Fu, Dongqi and Kang, Jian and Tong, Hanghang and He, Jingrui},
  journal={Advances in Neural Information Processing Systems},
  volume={37},
  pages={21342--21376},
  year={2024}
}

@article{qi2024robust,
  title={Robust neural contextual bandit against adversarial corruptions},
  author={Qi, Yunzhe and Ban, Yikun and Banerjee, Arindam and He, Jingrui},
  journal={Advances in Neural Information Processing Systems},
  volume={37},
  pages={19378--19446},
  year={2024}
}

@article{yang2026your,
  title={Your Group-Relative Advantage Is Biased},
  author={Yang, Fengkai and Chen, Zherui and Wang, Xiaohan and Lu, Xiaodong and Chai, Jiajun and Yin, Guojun and Lin, Wei and Ma, Shuai and Zhuang, Fuzhen and Wang, Deqing and others},
  journal={arXiv preprint arXiv:2601.08521},
  year={2026}
}

@article{lu2026contextual,
  title={Contextual Rollout Bandits for Reinforcement Learning with Verifiable Rewards},
  author={Lu, Xiaodong and Wang, Xiaohan and Chai, Jiajun and Yin, Guojun and Lin, Wei and Chen, Zhijun and Luo, Yu and Zhuang, Fuzhen and Ban, Yikun and Wang, Deqing},
  journal={arXiv preprint arXiv:2602.08499},
  year={2026}
}

@inproceedings{huang2025adaptive,
title={Adaptive Batch-Wise Sample Scheduling for Direct Preference Optimization},
author={Zixuan Huang and Yikun Ban and Lean Fu and Xiaojie Li and Zhongxiang Dai and Jianxin Li and deqing wang},
booktitle={The Thirty-ninth Annual Conference on Neural Information Processing Systems},
year={2025},
url={https://openreview.net/forum?id=8FN25PlktS}
}

@article{
qi2026bilevel,
title={Bi-level Hierarchical Neural Contextual Bandits for Online Recommendation},
author={Yunzhe Qi and Yao Zhou and Yikun Ban and Allan Stewart and Chuanwei Ruan and Jiachuan He and Shishir Kumar Prasad and Haixun Wang and Jingrui He},
journal={Transactions on Machine Learning Research},
issn={2835-8856},
year={2026},
url={https://openreview.net/forum?id=k3XsA75SGv},
note={J2C Certification}
}

\appendix


\section{Further Discussion}

\begin{figure}[h] 
     \includegraphics[width=0.47\columnwidth]{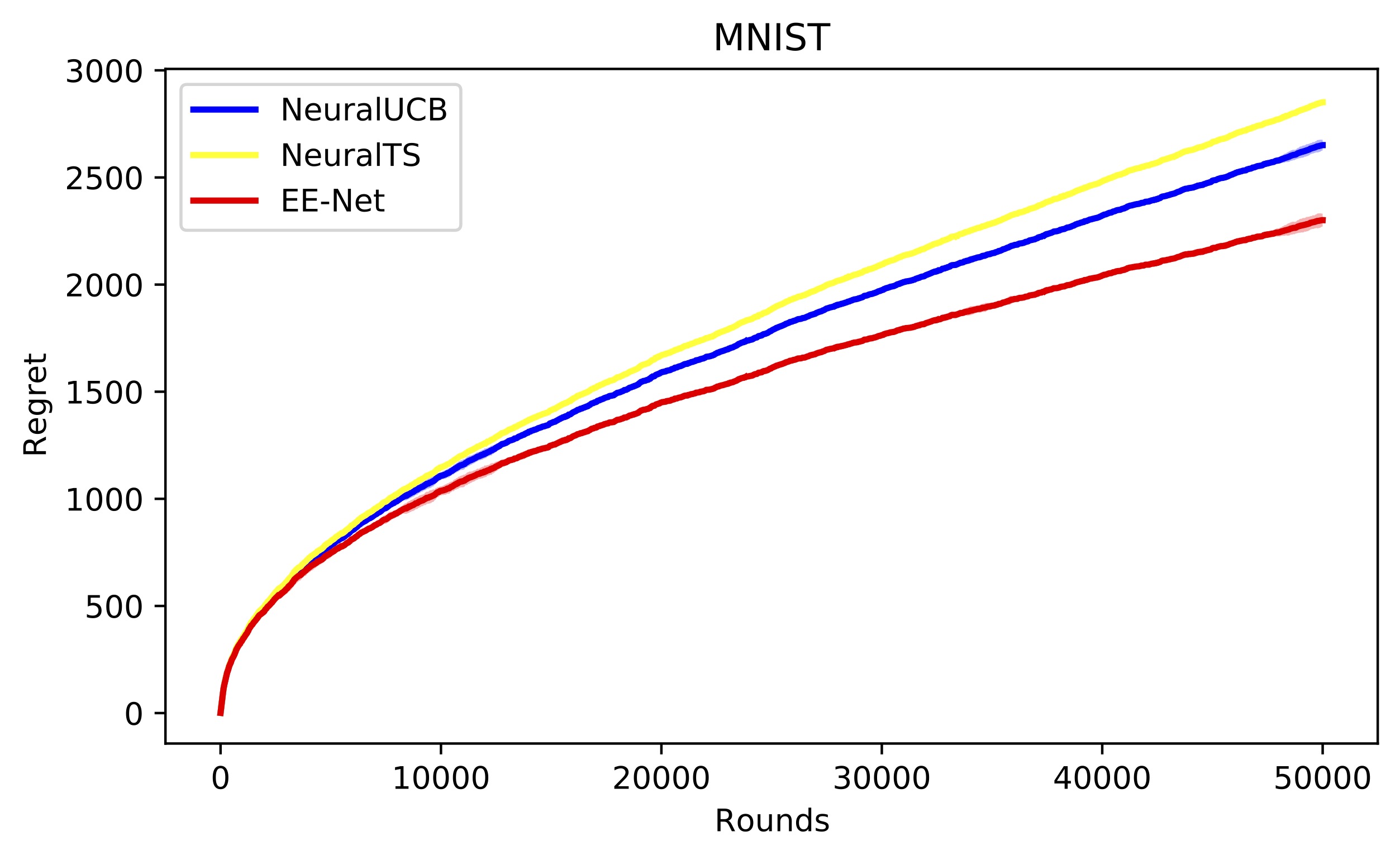}
    \includegraphics[width=0.47\columnwidth]{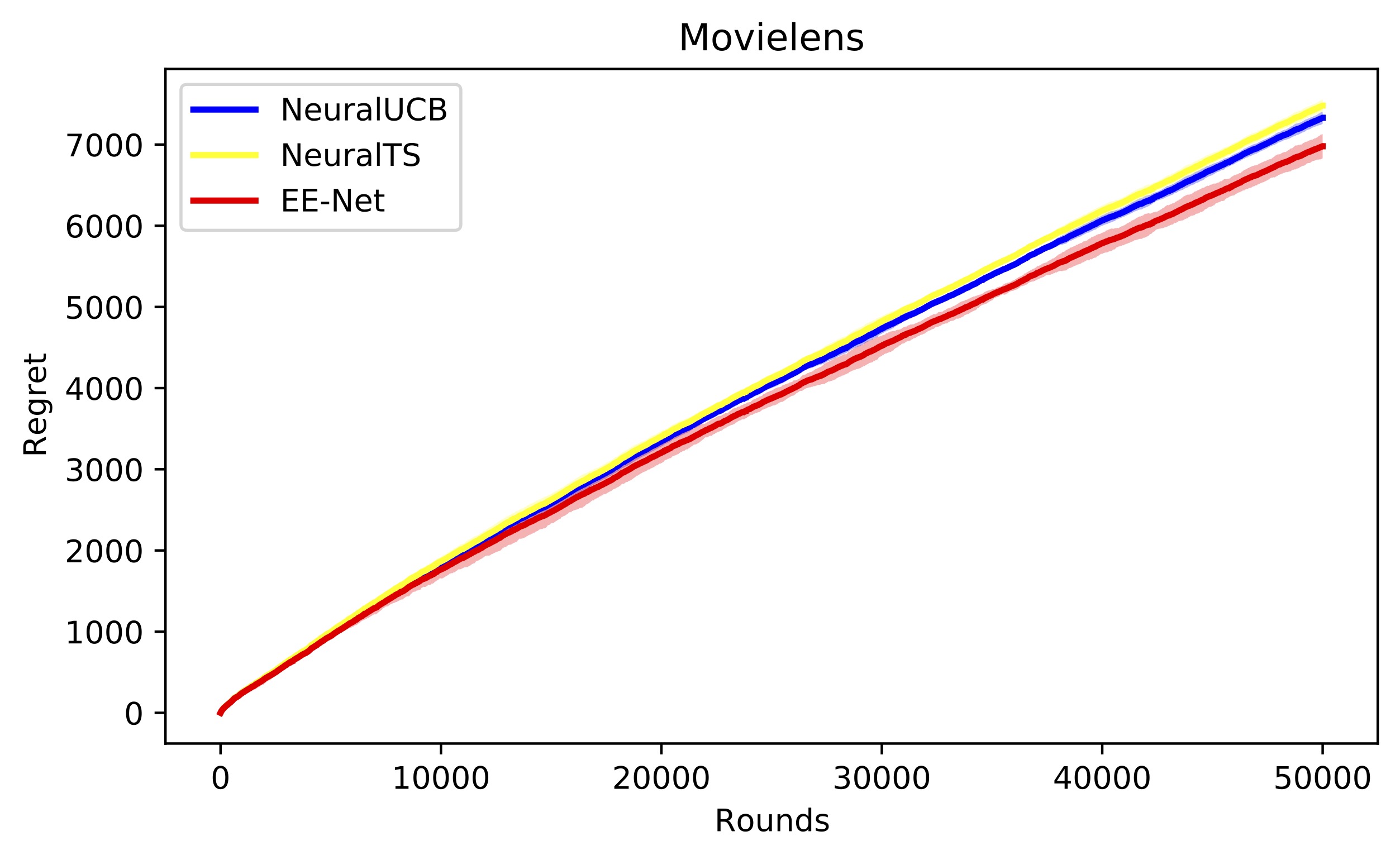}
    \centering
    \caption{Extended rounds on Movielens and MNIST Datasets}
       \label{fig:extendedrounds}
\end{figure}

Figure \ref{fig:extendedrounds} illustrates the extended rounds of regret compression on the Movielens and MNIST datasets, supporting our claim that the regret upper bound for \sysn is tighter than those for NeuralUCB and NeuralTS. As depicted in the figure, \sysn achieves the fastest convergence rate.

\subsection{Difference from Existing Works} \label{sec:indep}

As the linear bandits (Gentile et al., 2014; Li et al., 2016; Gentile et al., 2017; Li et al., 2019;) work in the Euclidean space and build the confidence ellipsoid for $\btheta^\ast$ (optimal parameters) based on the linear function $\mathbb{E}[r_{t,i} \mid \bx_{t,i}] = \langle \bx_{t,i}, \btheta^\ast \rangle $, their regret bounds contain $d$ because $\bx_{t,i} \in \mathbb{R}^{d}$. 
Similarly, neural bandits (Zhou et al., 2020; Zhang
et al., 2021) work in the RKHS 
and construct the confidence ellipsoid for $\btheta^\ast$ (neural parameters) according to the linear function $\mathbb{E}[r_{t, i} \mid \bx_{t,i}] = \langle \nabla_{\btheta_{0}}f(\bx_{t,i}; \btheta_0), \btheta^\ast - \btheta_0 \rangle$, 
where $\nabla_{\btheta_{0}}f(\bx_{t,i}; \btheta_0) \in \mathbb{R}^{p}$.
Their analysis is built on the NTK approximation, which is a linear approximation with respect to the gradient.  
Thus their regret bounds are affected by $\tilde{d}$ due to $\nabla_{\btheta_{0}}f(\bx_{t,i}; \btheta_0) \in \mathbb{R}^{p}$.
On the contrary, our analysis is based on the convergence error (regression)  and generalization bound of the neural networks. The convergence error term is controlled by the complexity term $\Psi^\ast$. 
And the generalization bound is the standard large-deviation bound, which depends on the number of data points (rounds).

\subsection{Upward and Downward Exploration}

\begin{table}[ht]
  \caption{}\label{tab:record}
  \centering
  \small
  \begin{tabular}{c|c|c} 
    \toprule
    Datasets & Upward Exploration & Downward Exploration \\
    \midrule
    Mnist & 76.3 \% & 23.7 \% \\
    Disin & 29.1 \% & 70.9 \% \\
    MovieLens & 58.6 \% & 41.4 \% \\
    Yelp & 55.3 \% & 44.7 \% \\
    \bottomrule
  \end{tabular}
  \vspace{-1em}
\end{table}

Table \ref{tab:record} shows the proportions of upward and downward explorations recorded over 10,000 rounds in real-world datasets.  In each round, we compared the reward estimated by the exploitation model $f_1$ with the received reward for each arm, determining the exploration direction.  The varying proportions across datasets suggest that identifying the exploration direction can provide more information and may be beneficial for balancing exploitation and exploration, beyond just considering exploration strength.

\subsection{Evaluation of Neural-Epsilon}

\begin{table}[h]
	\vspace{-1em}
	\caption{ Cumulative regret of Neural-$\epsilon$ variant.
 }\label{tab:epsilonplus}
	\vspace{1em}
	\centering
 \small
	\begin{tabular}{c|c|c|c|c}
		\toprule
		     & MNIST &  Disin & Movielens & Yelp \\
		\toprule   
                Neural-$\epsilon$ & 1126.8 $\pm$ 6   & 734.2 $\pm$ 31 & 1573.4  $\pm$ 26 &  5276.1 $\pm$ 27 \\
     Neural-$\epsilon^+$ & 1112.8 $\pm$ 8   & 724.2 $\pm$ 24 & 1578.4  $\pm$ 19 &  5162.4 $\pm$ 43 \\
		  \midrule  
            EE-Net &  842.3 $\pm$72 & 476.4 $\pm$ 23 & 1472.4 $\pm$ 5 &  4403.1  $\pm$ 13 \\
		\bottomrule
	\end{tabular}
	\vspace{-1em}
\end{table}

We also evaluate Neural-\(\epsilon\), where \(\epsilon\) is a decaying function of \(t\). The key intuition behind this approach is that exploration diminishes as more arms and rewards are observed. Specifically, we define \(\epsilon\) as \(\epsilon = \frac{\epsilon_0}{1 + \sqrt{t}}\), where \(\epsilon_0\) is a constant, and perform a grid search over \(\epsilon_0 \in \{0.01, 0.1, 0.2\}\). While this trade-off between exploration and exploitation allows Neural-\(\epsilon^+\) to achieve slightly better performance than Neural-\(\epsilon\), as shown in Table \ref{tab:epsilonplus}, it still falls short of EE-Net due to the unchanged random exploration mechanism.

\subsection{Motivation of Exploration Network} \label{sec:ucb}

\begin{table}[h]
	\caption{Selection Criterion Comparison ($\bx_t$: selected arm in round $t$).}\label{tab:1}
	\centering
	\begin{tabularx}{\textwidth}{c|X}
		\toprule
		   Methods   &   Selection Criterion  \\
		\toprule  
	       Neural-$\epsilon$  &  With probability $1-\epsilon$, $\bx_t = \arg \max_{\bx_{t,i}, i \in [n]} f_1(\bx_{t,i}; \btheta^1)$; Otherwise, select $ \bx_t$ randomly.\\
		  \midrule  
	       NeuralTS \citep{zhang2020neural} & For $\bx_{t, i}, \forall i \in [n]$, draw $\hat{r}_{t,i}$ from $\mathcal{N}(f_1(\bx_{t,i}; \btheta^1),  {\sigma_{t,i}}^2)$. Then, select $\bx_{t, \hat{i}}$, $\hat{i} = \arg \max_{i \in [n]} \hat{r}_{t,i}. $       \\
		\midrule  
	       NeuralUCB \citep{zhou2020neural} & $ \bx_t = \arg \max_{\bx_{t,i},  i \in [n]} \left( f_1(\bx_{t,i}; \btheta^1) +  \gamma_1  \| \nabla_{\btheta_t}f(\bx_{t, i}; \btheta_t) \|_{\mathbf{A}_t^{-1}} \right). $     \\
	   \midrule
	   \sysn &  $\bx_t = \arg \max_{\bx_{t,i} i \in [n]} \left(  f_1(\bx_{t,i}; \btheta^1) +   f_2 \left ( \nabla_{\btheta^1}f_1(\bx_{t,i}; \btheta^1); \btheta^2  \right)  \right)$. \\
		\bottomrule
	\end{tabularx}
	\vspace{-1em}
\end{table}

In this section, we list one gradient-based UCB from existing works \citep{ban2021multi, zhou2020neural}, which motivates our design of exploration network $f_2$.

\begin{lemma} \label{lemma:iucb}
(Lemma 5.2 in \citep{ban2021multi}). 
Given a set of context vectors $\{\bx_t \}_{t=1}^{T}$ and the corresponding rewards $\{r_t\}_{t=1}^{T} $ , $ \mathbb{E}(r_t) = h(\bxt)$ for any $\bx_t  \in \{\bx_t \}_{t=1}^{T}$. 
Let $f( \bx_t ; \btheta)$ be the $L$-layers fully-connected neural network where the width is $m$, the learning rate is $\eta$, the number of iterations of gradient descent is $K$.  Then, there exist positive constants $C_1, C_2,  S$,  such that if 
\[
\begin{aligned}
m &\geq  \text{poly} (T, n, L, \log (1/\delta) \cdot d \cdot e^{\sqrt{ \log 1/\delta}}), \ \ \eta =  \mathcal{O}( TmL +m \lambda )^{-1}, \ \   K \geq \widetilde{\mathcal{O}}(TL/\lambda),
\end{aligned} 
\]
then, with probability at least $1-\delta$, for any $\bx_{t,i}$, we have the following upper confidence bound:
\begin{equation}
\begin{aligned}
\left| h(\bx_{t,i})  - f(\bx_{t,i}; \btheta_t)     \right| \leq & \gamma_1  \| \nabla_{\btheta_t}f(\bx_{t, i}; \btheta_t)/\sqrt{m} \|_{\mathbf{A}_t^{-1}}  + \gamma_2  +  \gamma_1 \gamma_3  + \gamma_4,  \    
\end{aligned}
\end{equation}
where
\[ 
\begin{aligned}
& \gamma_1 (m, L) =  (\lambda + t \mathcal{O}(L)) \cdot ( (1 - \eta m \lambda)^{J/2} \sqrt{t/\lambda}) +1  \\
&\gamma_2(m,L, \delta) =    \| \nabla_{\btheta_t}f(\bx_{t, i}; \btheta_t)/\sqrt{m} \|_{\mathbf{A}_t^{' -1}} \cdot  \left( \sqrt{  \log \left(  \frac{\deter(\mathbf{A}_t')} { \deter(\lambda\mathbf{I}) }   \right)   - 2 \log  \delta } + \lambda^{1/2} S \right) \\
& \gamma_3 (m, L) = C_2 m^{-1/6} \sqrt{\log m}t^{1/6}\lambda^{-7/6}L^{7/2}, \ \  \gamma_4(m, L)= C_1 m^{-1/6}  \sqrt{ \log m }t^{2/3} \lambda^{-2/3} L^3 \\
& \mathbf{A}_t =  \lambda \mathbf{I} + \sum_{i=1}^{t}\nabla_{\btheta_t}f(\bx_{t, i}; \btheta_t) \nabla_{\btheta_t}f(\bx_{t, i}; \btheta_t)^\intercal /m , \ \,  \mathbf{A}_t' =  \lambda \mathbf{I} + \sum_{i=1}^{t} \nabla_{\btheta_0}f(\bx_{t, i}; \btheta_0) \nabla_{\btheta_0}f(\bx_{t, i}; \btheta_0)^\intercal /m .
\end{aligned}
\]
\end{lemma}
Note that $ \nabla_{\btheta_0}f(\bx_{t, i}; \btheta_0)$ is the gradient at initialization. Therefore, the above UCB can be represented as the following form for exploitation network $f_1$:  $|h(\bx_{t,i}) - f_1(\bx_{t,i}; \btheta^1_t)| \leq \kappa( \nabla_{\btheta_t^1}f_1(\bx_{t, i}; \btheta_t^1))$, where $\kappa$ is a mapping function.

Given an arm $x$, let $f_1(x)$ be the estimated reward and $h(x)$ be the expected reward. The exploration network $f_2$ in EE-Net is to learn $h(x) - f_1(x)$, i.e., the residual between expected reward and estimated reward, which is the ultimate goal of making exploration. There are advantages of using a network $f_2$ to learn $h(x) - f_1(x)$ in EE-Net, compared to giving a statistical upper bound for it such as NeuralUCB, \citep{ban2021multi}, and NeuralTS (in NeuralTS, the variance $\nu$ can be thought of as the upper bound).
For EE-Net, the approximation error for $h(x) - f_1(x)$ is caused by the genenalization error of the neural network (Lemma B.1. in the manuscript). In contrast, for NeuralUCB, \citep{ban2021multi}, and NeuralTS, the approximation error for $h(x) - f_1(x)$ includes three parts. 
The first part is caused by ridge regression. The second part of the approximation error is caused by the distance between ridge regression and Neural Tangent Kernel (NTK). The third part of the approximation error is caused by the distance between NTK and the network function. Because they use the upper bound to make selections, the errors inherently exist in their algorithms.

\begin{figure}[h] 
    \includegraphics[width=0.6\columnwidth]{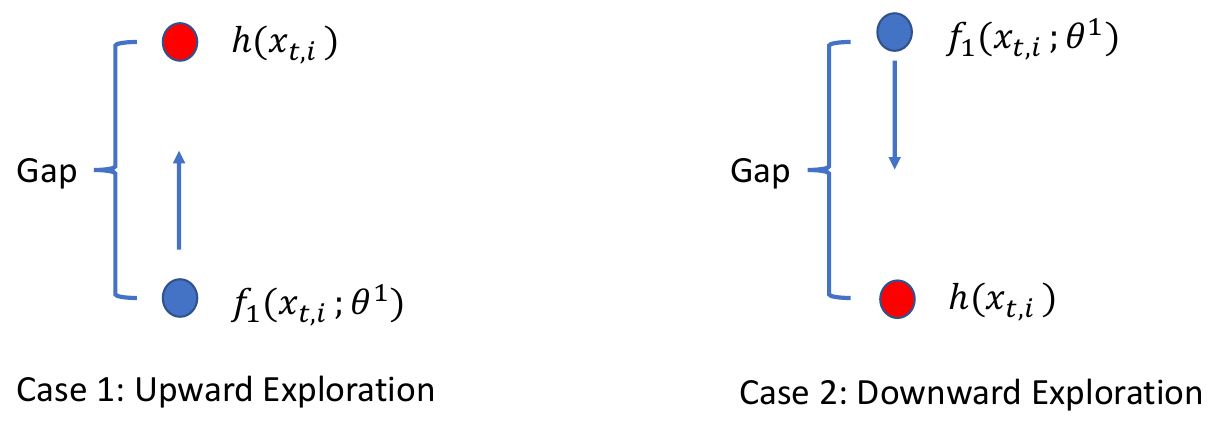}
    \centering
    \caption{ Two types of exploration: Upward exploration and Downward exploration. $f_1$ is the exploitation network (estimated reward) and $h$ is the expected reward.}
       \label{fig:twotype}
\end{figure}

The two types of exploration are described by Figure \ref{fig:twotype}.  When the estimated reward is larger than the expected reward, i.e., $h(x) - f_1(x) < 0$, we need to do the `downward exploration', i.e., lowering the exploration score of $x$ to reduce its chance of being explored; when $h(x) - f_1(x) > 0$, we should do the `upward exploration', i.e., raising the exploration score of $x$ to increase its chance of being explored. For EE-Net, $f_2$ is to learn $h(x) - f_1(x)$. When $h(x) - f_1(x) > 0$, $f_2(x)$ will also be positive to make the upward exploration. When  $h(x) - f_1(x) < 0$, $f_2(x)$ will be negative to make the downward exploration. 

\subsection{Discussion}

As \( f_2 \) is randomly initialized, it initially facilitates random exploration by estimating the potential gain of each unseen arm arbitrarily. When an arm is selected and its corresponding reward is observed, \( f_2 \) refines its potential gain estimation using the internal state (gradient) of \( f_1 \). In subsequent rounds, when \( f_1 \) encounters similar internal states (e.g., facing uncertainty in selecting arms), \( f_2 \) can provide more informed potential gain estimates to guide exploration.  
Over time, as the learner accumulates more internal state and potential gain pairs, \( f_2 \) gains a better understanding of various states of \( f_1 \), enabling more accurate potential gain estimation. Consequently, in each round, the combination of \( f_1 \) and \( f_2 \) computes an exploitation-exploration score for each arm, selecting the optimal one.  

In summary, during the initial phase, due to its random initialization, \( f_2 \) may produce inaccurate potential gain estimates, leading to suboptimal actions. However, it progressively improves by learning from these suboptimal decisions. As more observations are collected, \( f_2 \) refines its estimations, ultimately enhancing decision-making performance. Nevertheless, optimizing the initialization and training strategy of \( f_2 \) across different phases remains an interesting direction for future exploration.

\section{Proof of Theorem \ref{theo1}}

\subsection{Bounds for Generic Neural Networks}

In this section, we provide some base lemmas for the neural networks with respect to gradient or loss.
Let $\bx_t = \bx_{t, \hi}$.
Recall that $\call_t(\btheta) = (f(\bx_t; \btheta) - r_t)^2/2$.
Following \citep{allen2019convergence, cao2019generalization}, given an instance $\bx, \|\bx\| = 1$, we define the outputs of hidden layers of the neural network (Eq. (\ref{eq:structure})):
\[
\Gamma_0 = \bx,  \Gamma_l = \sigma(\bw_l \bh_{l-1}), l \in [L-1].
\]
Then, we define the binary diagonal matrix functioning as ReLU:
\[
\bD_l = \text{diag}( \mathbbm{1}\{(\bw_l \Gamma_{l-1})_1 \}, \dots, \mathbbm{1}\{(\bw_l \Gamma_{l-1})_m \} ), l \in [L-1],
\]
where $\mathbbm{1}$ is the indicator function :  $\mathbbm{1}(x) = 1$ if $x> 0$;  $\mathbbm{1}(x) = 0$, otherwise.
Accordingly, the neural network (Eq. (\ref{eq:structure})) is represented by 
\begin{equation}
f(\bx; \btheta^1 \ \text{or} \ \btheta^2) = \bw_L (\prod_{l=1}^{L-1} \bD_l \bw_l) \bx,
\end{equation}
and
\begin{equation}
\nabla_{\bw_l}f  = 
\begin{cases}
[\Gamma_{l-1}\bw_L (\prod_{\tau=l+1}^{L-1} \bD_\tau \bw_\tau)]^\top, l \in [L-1] \\
\Gamma_{L-1}^\top,   l = L .
\end{cases}
\end{equation}
Then, we have the following auxiliary lemmas.

\begin{lemma} \label{lemma:xi1}
 Suppose $m, \eta_1, \eta_2$ satisfy the conditions in Theorem \ref{theo1}.  With probability at least $1 -  \calo(TnL^2)\cdot \exp(-\Omega(m \omega^{2/3}L))$ over the random initialization, for all $t \in [T], i \in [n] $,  $\btheta$ satisfying $ \|\btheta - \btheta_0\|_2 \leq \omega$ with $ \omega \leq \calo(L^{-9/2} [\log m]^{-3})$, it holds uniformly that
\[
\begin{aligned}
&(1)  | f(\bx_{t,i}; \btheta)| \leq  \calo(1) \\
&(2)  \| \nabla_{\btheta}    f(\bx_{t,i}; \btheta) \|_2 \leq \calo(\sqrt{L}) \\
&(3) \|  \nabla_{\btheta} \call_t(\btheta)  \|_2 \leq  \calo(\sqrt{L})
\end{aligned}
\]
\end{lemma}

\begin{proof}
(1) is a simple application of Cauchy–Schwarz inequality.
\[
\begin{aligned}
| f(\bx_{t,i}; \btheta)| & = |  \bw_L (\prod_{l=1}^{L-1} \bD_l \bw_l) \bx_{t,i}  | \\
& \leq \underbrace{\|  \bw_L (\prod_{l=1}^{L-1} \bD_l \bw_l)  \|_2}_{I_1} \|\bx_{t,i} \|_2 \\
&  \leq \calo (1)
\end{aligned}
\]
where $I_1$ is based on the Lemma B.2 \citep{cao2019generalization}: $I_1 \leq \calo(1)$, and $\|\bx_{t,i}\|_2 = 1$.

For (2), it holds uniformly that
\[
 \| \nabla_{\btheta}    f(\bx_{t,i}; \btheta) \|_2  = \| \text{vec}( \nabla_{\bw_1}f )^\top,  \dots, \text{vec}( \nabla_{\bw_L}f )^\top\|_2 
 \leq \calo(\sqrt{L})
\]
where $\| \nabla_{\bw_l}f\|_F \leq \calo(1), l \in [L] $ is an application of Lemma B.3 \citep{cao2019generalization} by removing $\sqrt{m}$.

For (3), we have  $\| \nabla_{\btheta} \call_t(\btheta)\|_2 \leq | \call_t' | \cdot \|   \nabla_{\btheta}  f(\bx_{t,i}; \btheta) \|_2 \leq \calo(\sqrt{L})$.
\end{proof}

\begin{lemma} \label{lemma:functionntkbound}
Suppose $m, \eta_1, \eta_2$ satisfy the conditions in Theorem \ref{theo1}.  With probability at least $1 -  \calo(TnL^2)\cdot \exp(-\Omega(m \omega^{2/3}L))$ over the random initialization, for all $\| \bx \|_2 = 1$, $\btheta$ satisfying $ \|\btheta - \btheta_0\|_2,  \|\btheta' - \btheta_0\|_2 \leq \omega$ with $ \omega \leq \calo(L^{-9/2} [\log m]^{-3})$, it holds uniformly that
\[
| f(\bx; \btheta) - f(\bx; \btheta') -  \langle  \nabla_{\btheta'} f(\bx; \btheta'), \btheta - \btheta'    \rangle    | \leq \mathcal{O} (w^{1/3}L^2 \sqrt{ \log m}) \|\btheta - \btheta'\|_2.
\]
\end{lemma}

\begin{proof}
Based on Lemma 4.1 \citep{cao2019generalization}, it holds uniformly that
\[
|\sqrt{m} f(\bx; \btheta) - \sqrt{m} f(\bx; \btheta') -  \langle \sqrt{m}  \nabla_{\btheta'}f(\bx; \btheta'), \btheta - \btheta'    \rangle    | \leq \mathcal{O} (w^{1/3}L^2 \sqrt{m \log(m)}) \|\btheta - \btheta'\|_2,
\]
where $\sqrt{m}$ comes from the different scaling of neural network structure. Removing $\sqrt{m}$ completes the proof.
\end{proof}

\begin{lemma}  \label{lemma:differenceee}
Suppose $m, \eta_1, \eta_2$ satisfy the conditions in Theorem \ref{theo1}.  With probability at least $1 -  \calo(TnL^2)\cdot \exp(-\Omega(m \omega^{2/3}L))$ over the random initialization, for all $t \in [T], i \in [n] $,  $\btheta, \btheta' \in B(\btheta_0, \omega) $ with $ \omega \leq \calo(L^{-9/2} [\log m]^{-3})$, it holds uniformly that
\begin{equation}
\begin{aligned}
 \qquad  & |f(\bx_{t,i}; \btheta) - f(\bx_{t,i}; \btheta')|  \leq &   \calo(  \omega \sqrt{L})  +  \mathcal{O} (\omega^{4/3}L^2 \sqrt{ \log m})
 \end{aligned}
\end{equation}
\end{lemma}

\begin{proof}
Based on Lemma \ref{lemma:functionntkbound}, we have
\[
\begin{aligned}
 & |f(\bx_{t,i}; \btheta) - f(\bx_{t,i}; \btheta')|  \\
\overset{(a)}{\leq}  & |    \langle  \nabla_{\btheta'}f(\bx_{t,i}; \btheta'), \btheta - \btheta'    \rangle    | +  \mathcal{O} (\omega^{1/3}L^2 \sqrt{ \log m}) \|\btheta - \btheta'\|_2 \\
\overset{(b)}{\leq} & \| \nabla_{\btheta'}f(\bx_{t,i}; \btheta') \|_2 \cdot \| \btheta - \btheta' \|_2 + \mathcal{O} (\omega^{1/3}L^2 \sqrt{ \log m}) \|\btheta - \btheta'\|_2 \\
\overset{(c)}{\leq}  &\calo(\sqrt{L})  \| \btheta - \btheta' \|_2 +  \mathcal{O} (\omega^{1/3}L^2 \sqrt{ \log m}) \|\btheta - \btheta'\|_2 \\
\leq & \calo(\omega \sqrt{L}) + \calo(\omega^{4/3}L^2 \sqrt{\log m} ),
\end{aligned}
\]
where (a) is an application of Lemma \ref{lemma:functionntkbound}, (b) is based on the Cauchy–Schwarz inequality, and (c) is due to Lemma \ref{lemma:xi1}.
The proof is completed.
\end{proof}

\begin{lemma} [Almost Convexity of Loss]  \label{lemma:convexityofloss} 
Let $\call_t(\btheta) = (\sqrt{m}f(\bx_t; \btheta) - r_t)^2/2 $.  Suppose $m, \eta_1, \eta_2$ satisfy the conditions in Theorem \ref{theo1}.   With probability at least $1- \calo(TnL^2)\exp[-\Omega(m \omega^{2/3} L)] $ over randomness, for all $ t \in [T], i \in [n]$, and $\btheta, \btheta'$
satisfying $\| \btheta - \btheta_0 \|_2 \leq \omega$ and $\| \btheta' - \btheta_0 \|_2 \leq \omega$ with $\omega \leq \calo(L^{-6} [\log m]^{-3/2})$, it holds uniformly that
\[
 \call_t(\btheta')  \geq  \call_t(\btheta) + \langle \nabla_{\btheta}\call_t(\btheta),      \btheta' -  \btheta     \rangle  -  \epsilon.
\]
where $\epsilon =    \calo(\omega^{4/3}L^3 \sqrt{\log m}))  $

\end{lemma}

\begin{proof}
Let $\call_t'$ be the derivative of $\call_t$ with respective to $f(\bx_{t,i}; \btheta)$. Then, it holds that $ |  \call_t' | \leq \calo(1)$ based on Lemma \ref{lemma:xi1}.
Then, by convexity of $\call_t$, we have
\[
\begin{aligned}
&\call_t(\btheta') - \call_t(\btheta)  \\
\overset{(a)}{\geq} &  \call_t'  [  f(\bx_{t,i}; \btheta')  -  f(\bx_{t,i}; \btheta) ]  \\
\overset{(b)}{\geq}  & \call_t' \langle \nabla f(\bx_{t,i}; \btheta) ,  \btheta' -  \btheta\rangle  \\
 & - | \call_t'| \cdot  | f(\bx_{t,i}; \btheta')  -  f(\bx_{t,i}; \btheta) -   \langle \nabla f(\bx_{t,i}; \btheta) ,  \btheta' -  \btheta\rangle  |         \\
\geq & \langle \nabla_{\btheta}\call_t(\btheta ), \btheta' -  \btheta \rangle   -  | \call_t'| \cdot |  f(\bx_{t,i}; \btheta')  -  f(\bx_{t,i}; \btheta) -   \langle \nabla f(\bx_{t,i}; \btheta) ,  \btheta' -  \btheta\rangle  |  \\
 \overset{(c)}{\geq} &  \langle \nabla_{\btheta'}\call_t, \btheta' -  \btheta \rangle  -   \calo(\omega^{4/3}L^3 \sqrt{\log m}))  \\
 \geq & \langle \nabla_{\btheta'}\call_t, \btheta' -  \btheta \rangle - \epsilon
\end{aligned}
\]
where $(a)$ is due to the convexity of $\call_t$, $(b)$ is an application of triangle inequality,  and $(c)$ is the application of Lemma \ref{lemma:functionntkbound}. 
The proof is completed.
\end{proof}

\begin{lemma}[Trajectory Ball] \label{lemma:usertrajectoryball}
Suppose $m, \eta_1, \eta_2$ satisfy the conditions in Theorem \ref{theo1}. 
With probability at least $1- \calo(TnL^2)\exp[-\Omega(m \omega^{2/3} L)] $ over randomness of $\btheta_0$, for any $R > 0$, it holds uniformly  that
\[
\|\btheta_t^1 - \btheta_0\|_2 \leq \calo(R) \   \&  \ \|\btheta_t^2 - \btheta_0\|_2 \leq \calo(R),           t \in [T].
\]
\end{lemma}

\begin{proof}
Let $\omega  = \Omega(R)$.
The proof follows a simple induction. Obviously, $\btheta_0$ is in $B(\btheta_0, \omega)$.  Suppose that $\btheta_1, \btheta_2, \dots, \btheta_T \in \mathcal{B}(\btheta_0^2, \omega)$.
We have, for any $t \in [T]$, 
\begin{align*}
\|\btheta_T -\btheta_0\|_2 & \leq \sum_{t=1}^T \|\btheta_{t} -  \btheta_{t-1} \|_2 \leq \sum_{t=1}^T \eta \|\nabla \call_t(\btheta_{t-1})\| \leq  \sum_{t=1}^T \eta  \calo(\sqrt{L}) \\
&  =  \calo(TR^2 \sqrt{L} /\sqrt{m}) \leq \calo(R)
\end{align*}
The proof is completed.
\end{proof}

\begin{theorem} [Instance-dependent Loss Bound] \label{theo:instancelossbound}
Let $\call_t(\btheta) = (f(\bx_{t, \hi}; \btheta) - r_{t, \hi})^2/2$. 
Suppose $m, \eta_1, \eta_2$ satisfy the conditions in Theorem \ref{theo1}. 
With probability at least $1- \calo(TnL^2)\exp[-\Omega(m \omega^{2/3} L)] $ over randomness of $\btheta_0^2$, given any $R > 0$  it holds that
\begin{equation}
\sum_{t=1}^T \call_t(\btheta_{t-1}^2)  \leq  \sum_{t=1}^T  \call_t(\btheta^\ast) +  \calo(1) +  \frac{TLR^2}{\sqrt{m}} +  \calo( \frac{T R^{4/3} L^2\sqrt{\log m}}{m^{1/3}}).
\end{equation}
where $\btheta^\ast =  \arg \inf_{\btheta \in B(\btheta_0^2, R)}  \sum_{t=1}^T  \call_t(\btheta)$.
\end{theorem}

\begin{proof}
Let $\btheta' \in B(\btheta_0^2, R)$.
In round $t$, based on Lemma \ref{lemma:usertrajectoryball}, 
for all $t \in [T]$,  $\|\btheta_t - \btheta'\|_2 \leq \omega = \Omega(R)$,  it holds uniformly,
\begin{align*}
\call_t(\btheta_{t-1}^2) - \call_t(\btheta')  
\leq &  \langle \nabla \call_t(\btheta_{t-1}^2),  \btheta_{t-1}^2 - \btheta' \rangle   + \epsilon ,\\ 
\end{align*}
where $\epsilon = O(\omega^{4/3}L^2 \sqrt{\log m})$.

Therefore,  for all $t \in [T], \btheta' \in B(\btheta_0^2, R)$,  it holds uniformly
\begin{align*}
    \call_t(\btheta_{t-1}^2) - \call_t(\btheta')  \overset{(a)}{\leq} &  \frac{  \langle  \btheta_{t-1}^2 - \btheta_{t}^2 , \btheta_{t-1}^2 - \btheta'\rangle }{\eta_2}  + \epsilon ~\\
   \overset{(b)}{ = }  & \frac{ \| \btheta_{t-1}^2 - \btheta' \|_2^2 + \|\btheta_{t-1}^2 - \btheta_{t}^2\|_2^2 - \| \btheta_{t}^2 - \btheta'\|_2^2 }{2\eta_2}  + \epsilon ~\\
\overset{(c)}{\leq}& \frac{ \|\btheta_{t-1}^2  - \btheta'\|_2^2 - \|\btheta_{t}^2 - \btheta'\|_2^2  }{2 \eta_2}     + O(L\eta_2) + \epsilon\\
\end{align*}
where $(a)$ is because of the definition of gradient descent, $(b)$ is due to the fact $2 \langle A, B \rangle = \|A \|^2_F + \|B\|_F^2 - \|A  - B \|_F^2$, 
$(c)$ is by $ \|\btheta_{t-1}^2 - \btheta_{t}^2\|^2_2 = \| \eta_2 \nabla_{\btheta^2_{t-1}} \call_t(\btheta_{t-1}^2)\|^2_2  \leq \calo(\eta_2^2L)$.

Then,  for $T$ rounds,  we have
\begin{align*}
   & \sum_{t=1}^T \call_t(\btheta_{t-1}) -  \sum_{t=1}^T  \call_t(\btheta')  \\ 
   \overset{(a)}{\leq} & \frac{\|\btheta_0^2 - \btheta'\|_2^2}{2 \eta_2} + \sum_{t =2}^T \|\btheta_{t-1}^2 - \btheta' \|_2^2 ( \frac{1}{2 \eta_2} - \frac{1}{2 \eta_2}    )   + \sum_{t=1}^T L \eta_2 + T \epsilon    \\
   \leq & \frac{\|\btheta_0^2 - \btheta'\|_2^2}{2 \eta_2} + \sum_{t=1}^T L \eta_2 + T \epsilon    \\
   \overset{(b)}{\leq} & \calo(\frac{R^2}{ \sqrt{m} \eta_2}) + \sum_{t=1}^T L \eta_2 + T \epsilon  \\
   \overset{(c)}{\leq} & \calo(1) +  \frac{TLR^2}{\sqrt{m}} +  \calo( \frac{T R^{4/3} L^2\sqrt{\log m}}{m^{1/3}})
   \end{align*}
where $(a)$ is by simply discarding the last term and $(b)$ is because both $\btheta^2_0$ and $\btheta'$ are in the ball $B(\btheta^2_0, R)$, and $(c)$
is by $\eta_2 = \frac{R^2}{\sqrt{m}}$ and replacing $\epsilon$ with $\omega = \calo(R)$.  
The proof is completed.
\end{proof}

\subsection{Exploration Error Bound}

\newthetabound*

\begin{proof}
First,  
according to Lemma \ref{lemma:usertrajectoryball}, $\btheta^2_0, \dots, \btheta_{T-1}^2$ all are in $\mathcal{B}(\btheta_0, R)$. 
Then, according to Lemma \ref{lemma:xi1}, for any $\bx \in \bbr^d, \|\bx\|_2 = 1$, it holds uniformly $|f_1(\bx_{t, \hi}; \btheta^1_t)  + f_2( \phi(\bx_{t, \hi}); \btheta^2_t) - r_{t, \hi}| \leq \calo(1)$.

Then, for any $\tau \in [t]$, define
\begin{equation}
\begin{aligned}
V_{\tau} :=&\underset{ r_{\tau, \hi} }{\bbe} \left[ | f_2( \phi(\bx_{\tau, \hi}); \btheta^2_{\tau - 1}) -  (  r_{\tau, \hi} - f_1(\bx_{\tau, \hi}; \btheta^1_{\tau - 1})) | \right]  \\
& - | f_2( \phi(\bx_{\tau, \hi}); \btheta^2_{\tau - 1}) -  (  r_{\tau, \hi} - f_1(\bx_{\tau, \hi}; \btheta^1_{\tau - 1})) |
\end{aligned}
\end{equation}

Then, we have
\begin{equation}
\begin{aligned}
\bbe[V_{\tau}| F_{\tau - 1}]  =&\underset{r_{\tau, \hi} }{\bbe} \left[ | f_2( \phi(\bx_{\tau, \hi}); \btheta^2_{\tau - 1}) -  (  r_{\tau, \hi} - f_1(\bx_{\tau, \hi}; \btheta^1_{\tau - 1}))|\right] \\
& - \underset{r_{\tau, \hi} }{\bbe} \left[ | f_2( \phi(\bx_{\tau, \hi}); \btheta^2_{\tau - 1}) -  (  r_{\tau, \hi} - f_1(\bx_{\tau, \hi}; \btheta^1_{\tau - 1})) \right] \\
= & 0
\end{aligned}
\end{equation}
where $F_{\tau - 1}$ denotes the $\sigma$-algebra generated by the history $\mathcal{H}_{\tau -1}$. 

Therefore, the sequence $\{V_{\tau}\}_{\tau =1}^t$ is the martingale difference sequence.
Applying the Hoeffding-Azuma inequality, with probability at least $1-\delta$, we have
\begin{equation}
    \bbp \left[  \frac{1}{t}  \sum_{\tau=1}^t  V_{\tau}  -   \underbrace{ \frac{1}{t} \sum_{\tau=1}^t \underset{r_{i, \hi} }{\bbe}[ V_{\tau} | \mathbf{F}_{\tau-1} ] }_{I_1}   >  \sqrt{ \frac{2  \log (1/\delta)}{t}}  \right] \leq \delta   \\
\end{equation}    
As $I_1$ is equal to $0$, we have
\begin{equation}
\begin{aligned}
      &\frac{1}{t} \sum_{\tau=1}^t \underset{r_{\tau, \hi} }{\bbe} \left[ \left |  f_2( \phi(\bx_{\tau, \hi}); \btheta^2_{\tau - 1}) - (  r_{\tau, \hi} -  f_1(\bx_{\tau, \hi}; \btheta^1_{\tau - 1})) \right|   \right]  \\
 \leq  &  \underbrace{ \frac{1}{t}\sum_{\tau=1}^t  \left|f_2 ( \phi(\bx_{\tau, \hi}) ; \btheta_{\tau-1}^2)  - (r_{\tau, \hi} - f_1(\bx_{\tau, \hi}; \btheta_{\tau-1}^1))  \right| }_{I_3}  +   \sqrt{ \frac{2 \log (1/\delta) }{t}}   ~.
    \end{aligned}  
\label{eq:pppuper}
\end{equation}

For $I_3$, based on Theorem \ref{theo:instancelossbound}, for any $\btheta'$ satisfying $\| \btheta'  - \btheta^2_0 \|_2 \leq R /m^{1/4}$, with probability at least $1 -\delta$, we have
\begin{equation}
\begin{aligned}
I_3 &\leq 
\frac{1}{t}\sqrt{t}\sqrt{
\sum_{\tau =1}^t 
\left(     f_2 ( \phi(\bx_{\tau, \hi}) ; \btheta^2_{\tau-1} ) -   (r_{\tau, \hi}
- f_1(\bx_{\tau, \hi}; \btheta_{\tau-1}^1))
\right)^2}
\\
& \leq 
\frac{1}{t}\sqrt{t}\sqrt{
\sum_{\tau =1}^t 
\left(     f_2 ( \phi(\bx_{\tau, \hi}); \btheta') -   (r_{\tau, \hi}
- f_1(\bx_{\tau, \hi}; \btheta_{\tau-1}^1))
\right)^2}
+ \frac{ \calo(1)}{\sqrt{t}} 
 \\
& \overset{(a)}{\leq} \frac{ \sqrt{ \Psi(\btheta^2_0, R) } + \calo(1) } {\sqrt{t}}.
\end{aligned}
\end{equation}
where $(a)$ is based on the definition of instance-dependent complexity term. 
Combining the above inequalities together, with probability at least $1 - \delta$, we have 
\begin{equation}
\begin{aligned}
\frac{1}{t} \sum_{\tau=1}^t \underset{r_{\tau, \hi}}{\bbe} \left[   \left| f_2 (\phi(\bx_{\tau, \hi}); \btheta^2_{\tau-1}) - (  r_{\tau, \hi} -  f_1(\bx_{\tau, \hi}; \btheta^1_{\tau-1}) \right| \right] \\
\leq \frac{ \sqrt{ \Psi(\btheta^2_0, R) } + \calo(1) } {\sqrt{t}} +  \sqrt{ \frac{2  \log ( \calo(1)/\delta) }{t}}.
\end{aligned}
\end{equation}
The proof is completed.
\end{proof}

\begin{corollary}\label{corollary:main1} 
Suppose $m, \eta_1, \eta_2$ satisfy the conditions in Theorem \ref{theo1}.     For any $t \in [T]$,  let
\[
\iast = \arg \max_{ i \in [n]} \left[ h(\bx_{t,i})  \right],
\]
and $r_{t, \iast}$ is the corresponding reward, and denote the policy by $\pi^\ast$.
Let $\btheta^{1,\ast}_{t-1}, \btheta^{2,\ast}_{t-1}$ be the intermediate parameters trained by Algorithm \ref{alg:main} using the data select by $\pi^\ast$.
Then, with probability at least $(1- \delta)$  over the random of the initialization, for any $\delta \in (0, 1), R > 0$, it holds that
 \begin{equation}\label{eq:lemmamain}
 \begin{aligned}
 \frac{1}{t} \sum_{\tau=1}^t \underset{r_{\tau, \iast}}{\bbe} 
 & \left[   \left| f_2 ( \phi(\bx_{\tau, \iast}) ; \btheta_{\tau-1}^{2, \ast}) - \left(r_{\tau, \iast} - f_1(\bx_{\tau, \iast}; \btheta_{\tau-1}^{1, \ast}) \right)  \right| \mid  \pi^\ast, \mathcal{H}_{\tau-1}^\ast \right] \\
 & \qquad \qquad \leq 
 \frac{ \sqrt{ \Psi(\btheta^2_0, R) } + \calo(1) } {\sqrt{t}} +  \sqrt{ \frac{2  \log ( \calo(1)/\delta) }{t}},
 \end{aligned}
 \end{equation}
where  $\mathcal{H}_{\tau-1}^\ast = \{\bx_{\tau, \iast},  r_{\tau, \iast} \}_{\tau' = 1}^{\tau-1} $ represents the historical data produced by $\pi^\ast$ and the expectation is taken over the reward.
\end{corollary}

\begin{proof}
This a direct corollary of Lemma \ref{lemma:newthetabound}, given the optimal historical pairs $\{\bx_{\tau, \iast}, r_{\tau, \iast} \}_{\tau = 1}^{t-1}$ according to $\pi^\ast$.
For brevity, let $f_2( \phi(\bx);  \btheta_\tau^{2, \ast})$ represent $f_2\left( \nabla_{\btheta^{1, \ast}_\tau }f_1(\bx ; \btheta_\tau^{1, \ast}); \btheta_\tau^{2, \ast}\right)$. 

Suppose that, for each $\tau \in [t-1]$, 
$\btheta_\tau^{1, \ast}$ and $\btheta_\tau^{2, \ast}$ are the parameters training on $\{\bx_{\tau'}^\ast, r_{\tau'}^\ast \}_{\tau'=1}^\tau$ according to Algorithm \ref{alg:main} according to 
 $\pi^\ast$. 
Note that these pairs $\{\bx_{\tau'}^\ast, r_{\tau'}^\ast \}_{\tau'=1}^\tau$ are unknown to the algorithm we run, and the parameters $(\btheta_\tau^{1, \ast},\btheta_\tau^{2, \ast})$ are not estimated. However, for the analysis, it is sufficient to show that there exist such parameters so that the conditional expectation of the error can be bounded.

Then, we define 
\begin{equation} 
\begin{aligned}
V_{\tau} & := \underset{ r_{\tau, \iast} }{\bbe}
\left[ \left|f_2( \phi(\bx_{\tau, \iast}) ; \btheta_{\tau-1}^{2, \ast})  - \left(r_{\tau, \iast} - f_1(\bx_{\tau, \iast}; \btheta_{\tau-1}^{1, \ast})  \right)  \right|  \right] \\ 
 &\qquad \qquad - \left|f_2( \phi(\bx_{\tau, \iast}); \btheta_{\tau-1}^{2, \ast})  - \left(r_{\tau, \iast} - f_1(\bx_{\tau, \iast}; \btheta_{\tau-1}^{1, \ast})  \right) \right|~.
\end{aligned}
\end{equation}

Then, taking the expectation over reward, we have
 \begin{equation} \label{eq:vexp01}
 \begin{aligned}
 \bbe[  V_{\tau} | \mathbf{F}_{\tau-1}] & =  \underset{ r_{\tau, \iast}}{\bbe}
 \left[ \left|f_2\left(  \phi(\bx_{\tau, \iast}) ;  \btheta_{\tau-1}^{2, \ast}\right)  - \left(r_{\tau, \iast} - f_1(\bx_{\tau, \iast}; \btheta_{\tau-1}^{1, \ast})  \right)  \right|  \right]  \\
& \quad -   \underset{ r_{\tau, \iast} }{  \bbe}
\left[  \left|f_2 \left( \phi(\bx_{\tau, \iast}) ; \btheta_{\tau-1}^{2, \ast} \right)  - \left(r_{\tau, \iast} - f_1(\bx_{\tau, \iast}; \btheta_{\tau-1}^{1, \ast})  \right) \right| \mid \mathbf{F}_{\tau-1} \right] \\
& = 0~,
\end{aligned}
 \end{equation}
where $ \mathbf{F}_{\tau-1}$ denotes the $\sigma$-algebra generated by the history $\mathcal{H}_{\tau-1}^\ast = \{\bx_{\tau, \iast},  r_{\tau, \iast} \}_{\tau' = 1}^{\tau-1} $. 

Therefore,  $\{V_{\tau}\}_{\tau = 1}^t$ is a martingale difference sequence.
Similarly to Lemma \ref{lemma:newthetabound}, applying the Hoeffding-Azuma inequality to $V_{\tau}$, 
with probability  $1 - \delta$,  we have 

\begin{equation}
\begin{aligned}
 & \frac{1}{t} \sum_{\tau=1}^t  \underset{ r_{\tau, \iast}}{\bbe} \left[ \left|f_2 ( \phi(\bx_{\tau, \iast}) ; \btheta_{\tau-1}^{2, \ast} )  - \left(r_{\tau, \iast} - f_1(\bx_{\tau, \iast}; \btheta_{\tau-1}^{1, \ast})  \right) \right|  \right]  \\
 \leq &  \frac{1}{t}\sum_{\tau=1}^t \left|f_2 ( \phi(\bx_{\tau, \iast}) ; \btheta_{\tau-1}^{2, \ast} )  - \left(r_{\tau, \iast} - f_1(\bx_{\tau, \iast}; \btheta_{\tau-1}^{1, \ast}) \right)  \right|  + \sqrt{ \frac{2 \log (1/\delta) }{t}}  \\
 \leq &   
\frac{1}{t} \sqrt{t} \sqrt{ \sum_{\tau=1}^t \left(   f_2 ( \phi(\bx_{\tau, \iast}) ; \btheta^{2, \ast} ) - \left(r_{\tau, \iast} - f_1(\bx_{\tau, \iast}; \btheta_{\tau-1}^{1, \ast})   \right) \right)^2 }   + \sqrt{ \frac{2 \log (1/\delta) }{t}} \\
 \overset{(a)}{ \leq} &  \frac{1}{t} \sqrt{t} \sqrt{ \sum_{\tau=1}^t \left(   f_2 ( \phi(\bx_{\tau, \iast}) ; \btheta' ) - \left(r_{\tau, \iast} - f_1(\bx_{\tau, \iast}; \btheta_{\tau-1}^{1, \ast})   \right) \right)^2 }   + \frac{\calo(1)}{\sqrt{t}} + \sqrt{ \frac{2 \log (1/\delta) }{t}}  \\
\overset{(b)}{\leq} &  \frac{\sqrt{ \Psi(\btheta^2_0, R)} + \calo(1) }{\sqrt{t}} + \sqrt{ \frac{2 \log (1/\delta) }{t}} ,
\end{aligned}
\end{equation}
where $(a)$ is an application of Lemma \ref{theo:instancelossbound} for all $\btheta' \in B(\btheta_0^2, R)$ and $(b)$ is based on the definition of instance-dependent complexity term. 
Combining the above inequalities, the proof is complete.
\end{proof}

\subsection{Main Proof} \label{sec:the1p}

In this section, we provide the proof of Theorem \ref{theo1}.

\theomain*

\begin{proof}
For brevity, let $f(\bx; \btheta_{t-1}) = f_2 \left( \phi(\bx ) ) ; \btheta_{t-1}^2 \right) + f_1(\bx; \btheta_{t-1}^1).$
Then, the pseudo regret of round $t$ is given by
\begin{equation}
\begin{aligned}
R_t 
& = h(\bx_{t, \iast}) - h(\bx_{t,\hi})\\
& = \underset{ r_{t, i} , i \in [n]}{\bbe}[r_{t, \iast} - r_{t, \hi}]\\
& = \underset{ r_{t, i} , i \in [n]}{\bbe}[r_{t, \iast} - r_{t, \hi}]\\
&  =  \underset{ r_{t, i} , i \in [n]}{\bbe}[r_{t, \iast}  - f(\bx_{t,\hi}; \btheta_{t-1})   + f(\bx_{t,\hi}; \btheta_{t-1}) -r_{t, \hi} ]  \\
&  \leq \underset{ r_{t, i} , i \in [n]}{\bbe}[ \underbrace{  r_{t, \iast} - f(\bx_{t, \iast}; \btheta_{t-1})  +  f(\bx_{t,\hi}; \btheta_{t-1})   - f(\bx_{t,\hi}; \btheta_{t-1})  }_{I_1} +   f(\bx_{t,\hi}; \btheta_{t-1})  -r_{t, \hi} ]  \\
& =  \underset{ r_{t, i} , i \in [n]}{\bbe}[  r_{t, \iast} - f(\bx_{t, \iast}; \btheta_{t-1})  + f(\bx_{t,\hi}; \btheta_{t-1})   -r_{t, \hi}] \\
& \overset{(a)}{=} \underset{ r_{t, i} , i \in [n]}{\bbe}[  r_{t, \iast} -  f(\bx_{t, \iast}; \btheta_{t-1}^\ast)  +   f(\bx_{t, \iast}; \btheta_{t-1}^\ast)  -  f(\bx_{t, \iast}; \btheta_{t-1})  + f(\bx_{t,\hi}; \btheta_{t-1})  - r_{t, \hi} ]    \\
& \leq   \underset{ r_{t, i} , i \in [n]}{\bbe} 
  \left[  \left| f_2 \left( \phi(\bx_{t, \iast}); \btheta_{t-1}^{2, \ast} \right)   - \left(r_{t, \iast} - f_1(\bx_{t, \iast}; \btheta_{t-1}^{1, \ast})  \right) \right |  \right] \\
& \ \ \ +         
 \left| f_2 ( \phi(\bx_{t, \iast}) ; \btheta_{t-1}^{2, \ast} ) - f_2 (  \phi(\bx_{t, \iast}); \btheta^{2}_{t-1}) \right|   \\
& \ \ \  +      \left| f_1(\bx_{t, \iast}; \btheta_{t-1}^{1, \ast}) -  f_1(\bx_{t, \iast}; \btheta_{t-1}^{1}) \right|   \\
& \ \ \ +   \underset{ r_{t, i} , i \in [n]}{\bbe} 
  \left[   \left| f_2 ( \phi(\bx_{t, \hi})  ; \btheta_{t-1}^2 ) - \left(r_{t, \hi} - f_1(\bx_{t, \hi}; \btheta_{t-1}^1) \right) \right |  \right] \\ 
\end{aligned}
\end{equation}
where $I_1$ is because $f(\bx_{t,\hi}; \btheta_{t-1}) = \max_{i \in [n]}f(\bx_{t,i}; \btheta_{t-1})$ and $f(\bx_{t,\hi}; \btheta_{t-1}) - f(\bx_{t, \iast}; \btheta_{t-1}) \geq 0$ and $(a)$ introduces the intermediate parameters $\btheta_{t-1}^\ast = (\btheta_{t-1}^{1,\ast}, \btheta_{t-1}^{2,\ast})$ for analysis, which will be suitably chosen.

Therefore, we have
\begin{equation}
\begin{aligned}
\mathbf{R}_T = & \sum_{t = 1}^T R_t \\
\leq & \sum_{t = 1}^T  \underbrace{ \underset{ r_{t, i} , i \in [n]}{\bbe} 
  \left[  \left| f_2 ( \phi(\bx_{t, \iast}); \btheta_{t-1}^{2, \ast} ) - \left(r_{t, \iast} - f_1(\bx_{t, \iast}; \btheta_{t-1}^{1, \ast})  \right) \right | \right]}_{I_2} \\
&  +   \sum_{t = 1}^T \underbrace{        
 \left| f_2 \left( \phi(\bx_{t, \iast}) ; \btheta_{t-1}^{2, \ast} \right) - f_2 \left( \phi(\bx_{t, \iast}); \btheta^{1}_{t-1})) ; \btheta_{t-1}^{2} \right) \right| }_{I_3}  +   \sum_{t = 1}^T \underbrace{     \left| f_1(\bx_{t, \iast}; \btheta_{t-1}^{1, \ast}) -  f_1(\bx_{t, \iast}; \btheta_{t-1}^{1}) \right|  }_{I_4} \\
&   +   \sum_{t = 1}^T \underbrace{  \underset{ r_{t, i} , i \in [n]}{\bbe} 
  \left[  \left| f_2 \left( \phi(\bx_{t, \hi}) ; \btheta_{t-1}^2 \right)  - \left(r_{t, \hi} - f_1(\bx_{t, \hi}; \btheta_{t-1}^1)   \right)   \right |  \right]}_{I_5} \\ 
 \end{aligned}
\end{equation}

\begin{equation}
\begin{aligned}
 \leq & 2 \sum_{t=1}^T \Big( 
 \frac{\sqrt{ \Psi(\btheta^2_0, R)} + \calo(1) }{\sqrt{T}} + \sqrt{ \frac{2 \log ( \calo(1)/\delta) }{T}} \Big) \\
& +  2 \sum_{t=1}^T \Big( \calo(\sqrt{L}R ) + \calo(R^{4/3}L^2 \sqrt{\log m}/m^{1/3}) \Big)
 \\
\leq &  \sum_{t=1}^T \Big(  \frac{ 2\sqrt{ \Psi(\btheta^2_0, R)} + \calo(1) }{\sqrt{T}} + 2\sqrt{ \frac{2 \log ( \calo(1)/\delta) }{T}}   \Big)  \\
& + \calo(\sqrt{L}TR ) + \calo(TR^{4/3}L^2 \sqrt{\log m}/m^{1/3})  \\
\overset{(b)}{\leq} &  2\sqrt{ \Psi(\btheta^2_0, R) T} +  2\sqrt{ 2 \log (\calo(1)/\delta) T }  + \calo(1)
\end{aligned}
\end{equation}
$I_2$ and $I_5$ is based on Lemma \ref{lemma:newthetabound} and Corollary \ref{corollary:main1}
, i.e., $I_2, I_5 \leq  \frac{\sqrt{ \Psi(\btheta^2_0, R)} + \calo(1) }{\sqrt{T}} + \sqrt{ \frac{2 \log ( \calo(1)/\delta) }{T}}$ .
$I_3, I_4 \leq \calo(1)$ are based on Lemma \ref{lemma:differenceee}, respectively, because both $\btheta^2_{t-1}, \btheta^{2, \ast}_{t-1} \in B(\btheta_0^2, R)$ and $  \btheta^1_{t-1}, \btheta^{1, \ast}_{t-1} \in B(\btheta_0^1, R).$
$(b)$ is by the proper choice of $m$, i.e., when $m$ is large enough, we have $I_3, I_4 \leq \calo(1)$.

The proof is completed.
\end{proof}

\section{Connections with Neural Tangent Kernel}

\begin{lemma} [Lemma \ref{lemma:upperboundofStk} Restated]
Suppose $m$ satisfies the conditions in Theorem \ref{theo1}. With probability at least $1 - \delta$ over the initialization, there exists $\btheta'  \in  B(\btheta_0, \widetilde{\Omega}(T^{3/2}))$, such that
\begin{equation}
\begin{aligned}
 &\bbe[\Psi(\btheta^2_0, \widetilde{\Omega}(T^{3/2}))] \leq    \sum_{t=1}^{Tn} \bbe [ ( r_t -  f(\bx_t; \btheta'))^2/2 ] \\
\leq & \calo \left(\sqrt{ \widetilde{d} \log(1 + Tn) - 2 \log \delta  } + S + 1 \right)^2 \cdot \widetilde{d} \log (1+Tn).
\end{aligned}
\end{equation}
\end{lemma}

\begin{proof}
\[
\begin{aligned}
& \bbe [ \sum_{t=1}^{Tn} (r_t -  f(\bx_t; \btheta') )^2 ] = \sum_{t=1}^{Tn} ( h(\bx_t) - f(\bx_t; \btheta') )^2  \\
\overset{(a)}{\leq} &   \calo \left ( \sqrt{  \log \left(  \frac{\deter(\mathbf{A}_T)} { \deter(\mathbf{I}) }   \right)   - 2 \log  \delta }  +  S  + 1  \right)^2 \sum_{t=1}^{Tn}   \| \gx  \|_{\mathbf{A}_{T}^{-1}}^2 +  2Tn \cdot \calo \left(\frac{T^2 L^3 \sqrt{\log m}}{m^{1/3}} \right) \\
 \overset{(b)}{\leq} &  \calo \left(\sqrt{ \widetilde{d} \log(1 + Tn) - 2 \log \delta  } + S + 1 \right)^2 \cdot 
 \left( \widetilde{d} \log (1+Tn) + 1 \right) + \calo(1),
\end{aligned}
\]
where $(a)$ is based on Lemma \ref{lemma:bounfofsinglethetaprime} and $(b)$ is an application of Lemma 11 in \cite{2011improved} and Lemma \ref{lemma:detazero}, and $\calo(1)$ is induced by the choice of $m$.
By ignoring $\calo(1)$,  The proof is completed.
\end{proof}

\begin{definition} \label{def:ridge}
Given the context vectors $\{\bx_{\tau, \hi}\}_{\tau=1}^T$ and the rewards $\{ r_{\tau, \hi} \}_{\tau=1}^{T} $, then we define the estimation $\widehat{\btheta}_t$ via ridge regression:  
\[
\begin{aligned}
&\mathbf{A}_t = \mathbf{I} + \sum_{\tau=1}^{t} g(\bx_{\tau, \hi}; \btheta_0) g(\bx_{\tau, \hi}; \btheta_0)^\top \\
&\mathbf{b}_t = \sum_{\tau=1}^{t} r_{\tau, \hi} g(\bx_{\tau, \hi}; \btheta_0)  \\
&\widehat{\btheta}_t = \mathbf{A}^{-1}_t \mathbf{b}_t 
\end{aligned}
\]
\end{definition}

\begin{lemma}\label{lemma:bounfofsinglethetaprime}
Suppose $m$ satisfies the conditions in Theorem \ref{theo1}. With probability at least $1 - \delta$ over the initialization, there exists $\btheta'  \in  B(\btheta_0, \widetilde{\Omega}(T^{3/2}))$ for all $t \in [T]$, such that
\begin{equation}
\begin{aligned}
     &| \hx  - f(\bx_t; \btheta') | \\
\leq & \calo \left ( \sqrt{  \log \left(  \frac{\deter(\mathbf{A}_t)} { \deter(\mathbf{I}) }   \right)   - 2 \log  \delta }  +  S  + 1  \right) \| \gx  \|_{\mathbf{A}_{t}^{-1}} + \calo \left(\frac{T^2 L^3 \sqrt{\log m}}{m^{1/3}} \right)
\end{aligned}
\end{equation}
\end{lemma}

\begin{proof}
Given a set of context vectors $\{\bx_t\}_{t=1}^{Tn}$ with the ground-truth function $h$ and a fully-connected neural network $f$, we have
\[
\begin{aligned}
 &\left| \hx  - f(\bx_t; \btheta')     \right| \\
\leq &  \left  | \hx  - \langle \gx,  \htheta_t  \rangle \right|  + \left| f(\bx_t; \btheta')  -  \langle \gx, \htheta_t \rangle  \right|
\end{aligned}
\]
where $\btheta'$ is the estimation of ridge regression from Definition \ref{def:ridge}. Then, based on the Lemma \ref{lemma:existthetastar},
there exists $\bts \in \mathbf{R}^{P}$ such that $ h(\bx_t) =  \left \langle  g(\bx_t, \btheta_0), \bts \right \rangle$. Thus, we have
\[
\begin{aligned}
& \ \   \left  | \hx  - \langle \gx, \htheta_t \rangle \right| \\
  = & \left|  \left \langle  g(\bx_t, \btheta_0),   \bts \right \rangle   -   \left \langle  g(\bx_t, \btheta_0),  \htheta_t \right \rangle \right | \\
\leq   & \calo \left ( \sqrt{  \log \left(  \frac{\deter(\mathbf{A}_t)} { \deter(\mathbf{I}) }   \right)   - 2 \log  \delta }  +  S   \right) \| \gx  \|_{\mathbf{A}_{t}^{-1}}
\end{aligned}
\]
where the final inequality is based on the the Theorem 2 in \cite{2011improved}, with probability at least $1-\delta$, for any $t \in [T]$.

Second, we need to bound 
\[
\begin{aligned}
&\left| f(\bx_t; \btheta') -  \langle g(\bx_t; \btheta_0), \htheta_t \rangle  \right| \\
 \leq & \left |  f(\bx_t; \btheta') - \langle g(\bx_t; \btheta_0), \btheta' - \btheta_0 \rangle   \right|  \\
 &+    \left|     \langle  g(\bx_t; \btheta_0), \btheta' - \btheta_0 \rangle   - \langle  g(\bx_t; \btheta_0), \htheta_t \rangle \right|    
\end{aligned} 
\] 
To bound the above inequality,  we first bound
\[
\begin{aligned}
 & \left |  f(\bx_t; \btheta') - \langle g(\bx_t; \btheta_0), \btheta' - \btheta_0 \rangle   \right| \\
=& \left |  f(\bx_t; \btheta') - f(\mathbf{x}_t; \btheta_0)   - \langle g(\bx_t; \btheta_0), \btheta' - \btheta_0 \rangle   \right| \\
\leq  & \calo(\omega^{4/3} L^3 \sqrt{ \log m}) 
\end{aligned}
\] 
where  we  initialize $ f(\mathbf{x}_t; \btheta_0) = 0$ following \citep{zhou2020neural} and the inequality is derived by Lemma \ref{lemma:functionntkbound} with $\omega = \frac{\calo(t^{3/2})}{m^{1/4}}$. 
Next, we need to bound
\[ 
\begin{aligned}
 &|  \langle  g(\bx_t; \btheta_0), \btheta' - \btheta_0 \rangle - \langle  g(\bx_t; \btheta_0), \htheta_t \rangle | \\
 = & |\langle g(\bx_t; \btheta_0) ,     (\btheta' - \btheta_0 -  \htheta_t ) \rangle|  \\ 
\leq & \| g(\bx_t; \btheta_0)\|_{\mathbf{A}_t^{-1}} \cdot  \| \btheta' - \btheta_0  - \htheta_t\|_{\mathbf{A}_t} \\
\leq & \| g(\bx_t; \btheta_0)  \|_{\mathbf{A}_t^{-1}} \cdot  \|{\mathbf{A}_t} \|_2  \cdot  \| \btheta' - \btheta_0  - \htheta_t\|_2. \\
\end{aligned} 
\]
Due to the Lemma \ref{lemma:detazero} and Lemma \ref{lemma:2thetab}, we have
\[
\begin{aligned}
 & \|{\mathbf{A}_t} \|_2 \cdot   \| \btheta' - \btheta_0  - \htheta_t\|_2 \leq  (1 + t \mathcal{O}(L))   \cdot \frac{1}{1 + \calo(tL)}  =\calo(1).
 \end{aligned}
\] 
Finally, putting everything together, the proof is completed.
\end{proof}

\begin{definition}
\[
\begin{aligned}
&\mathbf{G}^{(0)} = \left[ g(\bx_{1, \hi}; \btheta_0), \dots,   g(\bx_{T, \hi}; \btheta_0)\right]  \in \bbr^{p \times T}   \\
&\mathbf{G}_0 = \left[ g(\bx_1; \btheta_0), \dots,   g(\bx_{Tn}; \btheta_0) \right]  \in \bbr^{p \times Tn} \\
&\mathbf{r}= (r_{1, \hi}, \cdots, r_{T, \hi}) \in \bbr^T
\end{aligned}
\]
$\mathbf{G}^{(0)}$ and $\mathbf{r}$ are formed by the selected contexts and observed rewards in $T$ rounds, $\mathbf{G}_0$ are formed by all the presented contexts.

Inspired by Lemma B.2 in \citep{zhou2020neural} , with $\eta = m^{-1/4}$ we define the auxiliary sequence following :
\[
\btheta_0 = \btheta^{(0)}, \ \ \btheta^{(j+1)}   = \btheta^{(j)} - \eta\left[ \bsg^{(0)} \left( [\bsg^{(0)}]^\top (\btheta^{(j)}  - \btheta_0) - \bsr \right)  + \lambda (\btheta^{(j)} - \btheta_0 )  \right] 
\]
\end{definition}

\begin{lemma} \label{lemma:existthetastar}
Suppose $m$ satisfies the conditions in Theorem \ref{theo1}. With probability at least $1 - \delta$ over the initialization, for any $t \in [T], i \in [K]$, the result uniformly holds:
\[
h(\bx_{t,i}) = \langle g(\bx_{t,i}; \btheta_0), \btheta^\ast - \btheta_0 \rangle.
\]

\end{lemma}
\begin{proof}
Based on Lemma \ref{lemma:boundgradientandNTK} with proper choice of $\epsilon$, we have
\[
\mathbf{G}^\top_0 \mathbf{G}_0 \succeq  \mathbf{H} - \|   \mathbf{G}^\top_0 \mathbf{G}_0 - \mathbf{H}    \|_F \mathbf{I} \succeq \mathbf{H} -\lambda_0 \mathbf{I}/2 \succeq \mathbf{H}/2 \succeq 0.
\] 
Define $\mathbf{h} = [h(\bx_1), \dots, h(\bx_{Tn})]$.
Suppose the singular value decomposition of $\mathbf{G}_0$ is $\mathbf{PAQ}^\top,   \mathbf{P} \in \bbr^{p \times Tn},  \mathbf{A} \in \bbr^{Tn \times Tn},  \mathbf{Q} \in \bbr^{Tn \times Tn}$, then, $\mathbf{A} \succeq 0$. 
Define  $\btheta^\ast = \btheta_0 + \mathbf{P} \mathbf{A}^{-1} \mathbf{Q}^\top \mathbf{h}$. Then, we have 
\[
\mathbf{G}^\top_0 (\btheta^\ast - \btheta_0) = \mathbf{QAP}^\top \mathbf{P}\mathbf{A}^{-1} \mathbf{Q}^{\top} \mathbf{h} = \mathbf{h}.
\]
which leads to 
\[
 \sum_{t=1}^{Tn}   ( h(\bx_{t}) - \langle g(\bx_{t}; \btheta_0), \btheta^\ast - \btheta_0 \rangle ) = 0.
\]
Therefore, the result holds:
\begin{equation}
\|     \btheta^\ast - \btheta_0       \|_2^2 = \mathbf{h}^\top \mathbf{QA}^{-2}\mathbf{Q}^\top \mathbf{h} =  \mathbf{h}^\top (\mathbf{G}^\top_0 \mathbf{G}_0)^{-1} \mathbf{h}  \leq  2 \mathbf{h}^\top\mathbf{H}^{-1} \mathbf{h} 
\end{equation}

\end{proof}

\begin{lemma} \label{lemma:2thetab}
There exist $\btheta'  \in  B(\btheta_0, \wcalo(T^{3/2}L + \sqrt{T}))$, such that,  with probability at least $1 -\delta$, the results hold:
\[
\begin{aligned}
& (1) \|         \btheta' - \btheta_0 \|_2 \leq  \frac{ \wcalo(T^{3/2}L + \sqrt{T })}{m^{1/4}} \\
& (2) \| \btheta' - \btheta_0 - \widehat{\btheta}_t  \|_2  \leq \frac{1}{1 +  \calo(TL)} \\
\end{aligned}
\]
\end{lemma}

\begin{proof}
\normalfont
The sequence of $\btheta^{(j)}$ is updated by using gradient descent on the loss function:
\[
\min_{\btheta} \mathcal{L}(\btheta) = \frac{1}{2}  \|[\bsg^{(0)}]^\top (\btheta - \btheta^{(0)} ) - \bsr  \|^2_2 + \frac{\lambda}{2} \|\btheta - \btheta^{(0)}\|_2^2 . 
\]

For any $j > 0$, the results hold: 
\[
\|   \bsg^{(0)}  \|_F  \leq \sqrt{T} \max_{t \in [T]} \|   g(\bx_t; \btheta_0)   \|_2 \leq \calo(\sqrt{TL}),
\]
where the last inequality is held by Lemma \ref{lemma:xi1}.
Finally, given the $j>0$, 
\begin{equation} \label{eq:boundofthetak}
\|  \btheta^{(j)} - \btheta^{(0)} \|_2 \leq \sum_{i=1}^{j} \eta \left[ \bsg^{(0)} \left( [\bsg^{(0)}]^\top (\btheta^{(i)}  - \btheta_0) - \bsr \right)  + \lambda (\btheta^{(i)} - \btheta_0 )  \right]  \leq \frac{\calo(j(TL\sqrt{T/\lambda} + \sqrt{T\lambda}))}{m^{1/4}}.
\end{equation}
For (2), by standard results of gradient descent on ridge regression, $\btheta^{(j)}$, and the optimum is $\btheta^{(0)} + \widehat{\btheta}_t $. Therefore, we have 
\[
\begin{aligned}
\| \btheta^{(j)} - \btheta^{(0)} - \widehat{\btheta}_t \|_2^2  & \leq \left[  1 -\eta \lambda\right]^j \frac{2}{\lambda} \left(  \mathcal{L}(\btheta^{(0)}) -  \mathcal{L}(\btheta^{(0)} + \widehat{\btheta}_t) \right) \\
 \leq &  \frac{2 (1 - \eta \lambda)^j}{\lambda}  \mathcal{L}(\btheta^{(0)}) \\
= & \frac{2 (1 - \eta m \lambda)^j}{ \lambda}  \frac{\|\bsr\|^2_2}{2} \\
\leq &\frac{T(1 - \eta \lambda)^j}{ \lambda}.
\end{aligned}
\]
By setting $\lambda = 1$ and $j  = \log ((T + \calo(T^2L))^{-1})/ \log (1 - m^{- 1/4})$, we have $ \| \btheta^{(j)} - \btheta_0 - \widehat{\btheta}_t \|_2^2 \leq \frac{1}{1 +  \calo(TL)} $.
Replacing $k$ and $\lambda$ in  \eqref{eq:boundofthetak} finishes the proof.
\end{proof}

\begin{lemma} \label{lemma:detazero}
Suppose $m$ satisfies the conditions in Theorem \ref{theo1}. With probability at least $1 - \delta$ over the initialization, the result holds:
\[
\begin{aligned}
\|  \mathbf{A}_T \|_2 &\leq 1  + \mathcal{O}(TL), \\
\log \frac{\det \mathbf{A}_T}{ \det \mathbf{I}} &\leq \widetilde{d} \log(1 + Tn) + 1.
\end{aligned}
\]
\end{lemma}

\begin{proof}
Based on the Lemma \ref{lemma:xi1}, for any $t \in  [T]$, 
$ \|g(\bx_t; \btheta_0) \|_2 \leq \mathcal{O}(\sqrt{L})$.
Then, for the first item:
\[
\begin{aligned}
&\|  \mathbf{A}_T \|_2  =  \|   \mathbf{I} + \sum_{t=1}^{T} g(\bx_{t, \hi}; \btheta_0) g(\bx_{t,\hi}; \btheta_0)^\top \|_2 \\
& \leq   \| \mathbf{I} \|_2 + \| \sum_{t=1}^{T} g(\bx_{t, \hi}; \btheta_0) g(\bx_{t, \hi}; \btheta_0)^\top \|_2  \\ 
&\leq  1  +  \sum_{t=1}^{T} \| g(\bx_{t, \hi}; \btheta_0) \|_2^2  \leq 1  + \mathcal{O}(TL).
\end{aligned}
\]
Next, we have
\[
\log \frac{\deter(\mathbf{A}_T) }{\deter(   \mathbf{I})} \leq \log \deter(\mathbf{I} + \sum_{t=1}^{Tn} g(\bx_t; \btheta_0) g(\bx_t; \btheta_0)^\top  ) = \deter( \mathbf{I} + \mathbf{G}_0 \mathbf{G}_0^{\top}) 
\]

Then, we have 
\[
\begin{aligned}
 &\log \det(\mathbf{I} + \mathbf{G}_0 \mathbf{G}^{\top}_0  ) \\
& = \log \deter ( \mathbf{I} + \mathbf{H}   +  (\mathbf{G}_0 \mathbf{G}^{\top}_0 - \mathbf{H})    ) \\
& \leq   \log \deter ( \mathbf{I} +  \mathbf{H}  ) + \langle ( \mathbf{I} +  \mathbf{H}   )^{-1},   (\mathbf{G}_0 \mathbf{G}^{\top}_0 - \mathbf{H})    \rangle \\
& \leq \log \deter(  \mathbf{I} +  \mathbf{H}  ) +  \| ( \mathbf{I} +  \mathbf{H}   )^{-1}    \|_{F} \|  \mathbf{G}_0 \mathbf{G}^{\top}_0 - \mathbf{H}  \|_F    \\
& \leq  \log \deter(  \mathbf{I} +  \mathbf{H}  ) +  \sqrt{T}\|  \mathbf{G}_0 \mathbf{G}^{\top}_0 - \mathbf{H}  \|_F   \\
&\leq   \log \deter(  \mathbf{I} +  \mathbf{H}  ) +  1\\
&= \widetilde{d} \log ( 1 + Tn )+  1. 
\end{aligned}
\]
The first inequality is because the concavity of $\log \deter$ ; The third inequality is due to $  \| ( \mathbf{I} +  \mathbf{H} \lambda )^{-1} \|_{F} \leq  \| \mathbf{I}^{-1} \|_{F}  \leq \sqrt{T}$; The last inequality is because of the choice the $m$, based on Lemma \ref{lemma:boundgradientandNTK}; The last equality is because  of the Definition of $\hd$.
The proof is completed.
\end{proof}

\begin{lemma} \label{lemma:boundgradientandNTK}
For any $\delta \in (0, 1)$, if $m = \Omega \left( \frac{L^6 \log(TnL/\delta)}{(\epsilon/Tn)^4} \right)$, then with probability at least $1 - \delta$, the results hold:
\[
\|   \mathbf{G}_0 \mathbf{G}^{\top}_0 - \mathbf{H}  \|_F \leq \epsilon.
\]
\end{lemma}
\begin{proof}
This is an application of Lemma B.1 in \citep{zhou2020neural} by properly setting $\epsilon$.
\end{proof}

\end{document}